\documentclass{article}
\pdfoutput=1
\PassOptionsToPackage{numbers, compress}{natbib}

\usepackage[preprint]{neurips_2021}

\usepackage[utf8]{inputenc} 
\usepackage[T1]{fontenc}    
\usepackage{url}            
\usepackage{booktabs}       
\usepackage{amsfonts}       
\usepackage{nicefrac}       
\usepackage{microtype}      
\usepackage{xcolor}         

\usepackage{amssymb}
\usepackage{amsmath}
\usepackage{mathtools}
\usepackage{caption}
\usepackage{subcaption}
\usepackage{rotating}
\usepackage[raggedright]{sidecap}
\usepackage{graphicx}
\usepackage{algorithm} 

\newcommand{\define}{\triangleq}

\newcommand{\Hessian}{{\mathbf{H}}}
\newcommand{\noise}{\xi}
\newcommand{\variance}{\sigma}
\newcommand{\prob}{\mathcal{N}}
\newcommand{\bw}{W}
\newcommand{\btheta}{{\sigma^2}}
\newcommand{\bG}{\Upsilon}
\newcommand{\br}{\eta}

\newcommand{\hyper}{\eta} 
\newcommand{\Hyper}{\boldsymbol{\nu}}

\newtheorem{prop}{Proposition}
\newtheorem{pf}{Proof}
\newcommand{\bN}{\mathcal{N}}
\newcommand{\by}{Y}
\newcommand{\bx}{X}
\newcommand{\data}{K}
\newcommand{\R}{\mathbb{R}}

\DeclareMathOperator{\diag}{diag}
\DeclareMathOperator*{\argmax}{argmax}
\newcommand{\argmin}{\text{arg}\min}
\newcommand{\mean}{\mathbf{\mu}}

\newcommand{\Gradient}{\mathbf{g}}

\newcommand{\mask}{\zeta}
\newcommand{\energy}{\frac{1}{2\variance^2}\sum_{k=1}^{\data}(Y_{k}-\text{Net}(\bx_k,\bw))^2}

\newcommand{\bgamma}{\bG}
\newcommand{\balpha}{\boldsymbol{\alpha}}
\newcommand{\quadratic}{{\frac{1}{2}(\bw-\bw^*)^{\top}\Hessian(\bw^*, \btheta)(\bw-\bw^*)+(\bw-\bw^*)^{\top}\mathbf{G}(\bw^*, \btheta)}}

\title{Bayesian Learning to Discover \\Mathematical Operations in Governing Equations \\of Dynamic Systems}

\author{
Hongpeng Zhou\\
  Department of Cognitive Robotics\\
  Delft university of technology\\
  Mekelweg 2, 2628 CD Delft\\
  \texttt{h.zhou-3@tudelft.nl} \\
   \And
   Wei Pan \\
	Department of Cognitive Robotics\\
	Delft university of technology\\
	Mekelweg 2, 2628 CD Delft\\
	\texttt{Wei.Pan@tudelft.nl} \\
}

\begin{document}

\maketitle
\begin{abstract}
	Discovering governing equations from data is critical for diverse scientific disciplines as they can provide insights into the underlying phenomenon of dynamic systems. This work presents a new representation for governing equations by designing the Mathematical Operation Network (MathONet) with a deep neural network-like hierarchical structure. Specifically, the MathONet is stacked by several layers of unary operations (e.g. \texttt{sin}, \texttt{cos}, \texttt{log}) and binary operations (e.g. $+,-,\times$), respectively. An initialized MathONet is typically regarded as a super-graph with a redundant structure, a sub-graph of which can yield the governing equation. We develop a sparse group Bayesian learning algorithm to extract the sub-graph by employing structurally constructed priors over the redundant mathematical operations. By demonstrating on the chaotic Lorenz system, Lotka-Volterra system, and Kolmogorov–Petrovsky–Piskunov system, the proposed method can discover the ordinary differential equations (ODEs) and partial differential equations (PDEs) from the observations given limited mathematical operations, without any prior knowledge on possible expressions of the ODEs and PDEs.
\end{abstract}

\section{Introduction}
\label{sec:intro}
Accurate governing equations (e.g., ordinary differential equations (ODEs) or partial differential equations (PDEs)) can facilitate the study on robust prediction, system control, stability analysis, and increase the interpretability of a physical process~\citep{rudy2019deep,rasmuson2014mathematical}. The first principle method is dominating to obtain the governing equation, for example, one for pendulums using Newton's second law of motion to pertains to the relation between the acceleration of an object and the external forces acting on it. Nevertheless, slow modeling process~\citep{lin1988mathematics} and lack of domain knowledge~\citep{wollkind2017comprehensive} of the first principle method motivate the shift to a data-driven approach.

A seminal breakthrough from~\citep{Schmidt2009science} applied symbolic regression~\citep{Goldberg1989Genetic} to determine the underlying structure and parameters of time-invariant non-linear dynamic systems. Typically, the symbolic regression method~\cite{Schmidt2009science} is too computationally expensive and is prone to overfitting. \citep{Brunton3932,hoffmann2019reactive,quade2018sparse,rudy2017data} used sparse regression techniques to determine the fewest terms in the dynamic governing equations to represent the data accurately. The sparse regression approaches of~\cite{Brunton3932} cannot avoid the nontrivial task of choosing appropriate sets of basis functions and cannot build the model with more unary functions (e.g. such as \texttt{sin}, \texttt{cos}, \texttt{log}).
More recently, physical-based machine learning methods arouse much interest, which focuses on combining physical-based modeling and machine learning methods~\citep{willard2020integrating,von2019informed}.
\citep{raissi2019physics} proposed the physics-informed neural networks (PINNs) to discover both the solution and structure of PDEs by encoding prior knowledge in the cost functions.
The physics-informed neural networks~\citep{raissi2019physics, rackauckas2020universal} require prior knowledge and belong to the grey-box approximator, which lacks complete interpretability.

This paper is motivated by both symbolic regression and sparse regression to discover governing equations of dynamic systems. Similar to symbolic regression, we start from some basic mathematical operations, i.e., unary (e.g. \texttt{sin},\texttt{cos},\texttt{log}) and binary operations (e.g. $+, \times$), instead of a dictionary of predefined equation terms. Instead of exploring the best fitness by trying possible combinations of operations as much as possible, we cast the discovery problem as a sparse optimization problem. Specifically, the optimization is nothing different than training a sparse deep neural network (DNN), i.e., DNN compression problem~\citep{lee2018snip, han2015deep, wen2016learning}, which can be solved using sparse group Lasso type of algorithms. The key idea is to reformulate the governing equation composed of unary and binary operations into a \emph{augmented hierarchical structure} similar to DNN. In Fig.~\ref{fig:aritho_net}(a) and (c), a tutorial is given to show how an expression of $k_4 \texttt{sin}({k_1}y+{k_2}y^2)+{k_5}\texttt{cos}({k_3}x^3)$ can be decomposed into a DNN-like model, termed as Mathematical Operation Network (MathONet). The model in Fig.~\ref{fig:aritho_net}(c) can be augmented into a over-parametrized network (see Fig.~\ref{fig:aritho_net}(d)) by adding extra mathematical operations. In MathONet, the operations are similar to the activation function in DNN; and the weights of the connection between two operations are expected to be zeros if the corresponding operations are redundant. To this end, the true underlying governing equations can be seen as a sub-graph of an over-parameterized super-graph of the MathONet. Essentially, the governing equation discovery problem is equivalent to the compression problem of MathONet with structural and non-structural sparsities.

Among various DNN compression techniques~\citep{lecun1990optimal,hassibi1993optimal,han2015deep}, the structured sparsity learning method in~\cite{wen2016learning} is the most related algorithm for our approach and can be readily applied. This method is essentially a sparse group Lasso type of algorithm~\citep{simon2013sparse} where $\|{\bw}\|_{\ell_1}$ and $\|{\bw_g}\|_{\ell_2}$ are added on top of the conventional loss function of data as regularizer to promote sparsity of the network weight $\bw$. Our practical implementation found that the solutions are always non-sparse compared to the true underlying governing equations, even when the hyper-parameters are extensively tuned (more details can be found in Fig.~\ref{fig:lorenz_weight_sparsity_loss_change_model_z} in Sec.~\ref{sebsec:Lorenz System}, Fig.~\ref{fig:Lotka_Volterra_weight_sparsity_loss_change} in Appendix.Sec.~\ref{appsubsubsec:Lotka-Volterra_result} and Fig.~\ref{fig:fisher_weight_sparsity_loss_change} in Appendix.Sec.~\ref{appsubsubsec:Fisher-KPP_result}). This motivates us to seek a Bayesian learning solution that can potentially result in sparser solutions. The Bayesian learning approach can incorporate the structural and non-structural sparsity as priors in a principled manner. The algorithm is demonstrated on chaotic Lorenz system, Lotka-Volterra system, and Kolmogorov–Petrovsky–Piskunov system. The proposed method can discover the ordinary differential equations (ODEs) and partial differential equations (PDEs) from the observations given limited mathematical operations, without any prior knowledge on possible expressions of the ODEs and PDEs. Overall, the contribution of paper has two folds:

\textbf{Mathematical operation network design:}
The governing equations of dynamic systems (ODEs or PDEs) characterized by basic mathematical operations, i.e., unary and binary ones, can be represented as a DNN-like hierarchical structure, termed MathONet. The governing equation discovery problem can be treated as a sub-graph search problem from an over-parameterized MathONet graph with redundant mathematical operations.

\textbf{Bayesian learning discovery algorithm :} 
A Bayesian learning alternative of the sparse group Lasso type of algorithms is proposed to discover a more sparse solution, the mathematical operations in governing equations without too much hyperparameter tuning. The algorithm was demonstrated on some well-known dynamics systems in physics and ecology, for which the governing equations are learned from scratch given basic mathematical operations without any prior knowledge on the format of the underlying governing equations.   

\vspace{-0.3cm}
\section{Related Works}
\label{sec:related work}

\vspace{-0.2cm}
\paragraph{System identification}
Data-driven discovery of the governing equations has become an active research area for many years~\citep{roscher2020explainable,carleo2019machine}. With provided heuristics and expert guidance, several pioneering works started from rediscovering known governing equations in specific disciplines (e.g., Proust’s law in chemistry~\citep{langley1981data}, ideal gas law~\citep{Langley1983}) with simulated data~\citep{lenat1983role}. Further work was also implemented for equation discovery of ecological applications with real collected dataset~\cite{dvzeroski1999equation}. Another typical classical method is system identification, which aims to obtain an approximated mathematical model by identifying the model parameters~\citep{ljung1999system}. However, since the typical prerequisite for system identification is that the model structure is known, these methods are impractical for the domains without known mathematical laws(e.g., neuroscience, cell biology, finance, epidemiology)~\citep{dvzeroski1999equation, willard2020integrating}. To improve generalization, the method of discovering both structure and coefficients of the governing equations becomes a key research direction.

\vspace{-0.2cm}
\paragraph{Symbolic regression}
Symbolic regression can be used to discover mathematical operations in governing equations without prior domain knowledge~\citep{Schmidt2009science, Goldberg1989Genetic}. In \cite{udrescu2020symbolic}, symbolic regression is applied for unsupervised learning of motion equations of the object from a distorted unlabelled video. Other successful applications include automated refinement and reverse engineering for metabolic networks~\citep{bongard2007automated, Schmidt_2011}, synchronization control of oscillator networks~\citep{gout2018synchronization}, motion prediction of harmonic oscillator~\citep{PhysRevE.94.012214} and constructing smooth value functions for reinforcement learning~\citep{kubalik2019symbolic}, etc.
The symbolic regression method can learn the mathematical expression by searching over a space of basic arithmetic operations and self-evolved by genetic algorithms~\citep{Schmidt2009science,Goldberg1989Genetic}.
However, it is too computationally expensive to scale well to high-dimensional systems and large datasets.

\vspace{-0.2cm}
\paragraph{Sparse regression}
The sparse regression is a promising technique that can determine (almost) the fewest equation terms to describe a system. In \cite{Brunton3932}, sparse identification of non-linear dynamics (SINDy) algorithm is proposed to identify the governing equations as a linear combination of basis functions selected from a pre-built function dictionary. The SINDy method has achieved success on many benchmarks, e.g., fluid dynamics~\citep{Brunton3932}, chemical kinetics~\cite{hoffmann2019reactive}. It was also expanded to address the model recovery of dynamic systems following abrupt changes~\citep{quade2018sparse} and the discovery of partial differential equations~\cite{rudy2017data}. These works show that the sparse regression methods provide an effective manner to identify the governing equations. However, these approaches also suffer from the non-trivial task of choosing appropriate basis functions, limiting their capacity for more general applications.

Both the symbolic and sparse regression techniques explore the governing equations within an ample space of possibly non-linear mathematical terms. 
The comparisons and discussions between them can be found in a recent work~\citep{SUBRAMANIAN2021100014}, which identifies dynamical equations of a distillation column.

\vspace{-0.2cm}
\section{MathONet Design}
\label{sec:network}

\vspace{-0.2cm}
\subsection{Motivation}
\label{sunsec:design motvation}

\begin{figure}[ht]
	\centering
		\includegraphics[width=\textwidth]{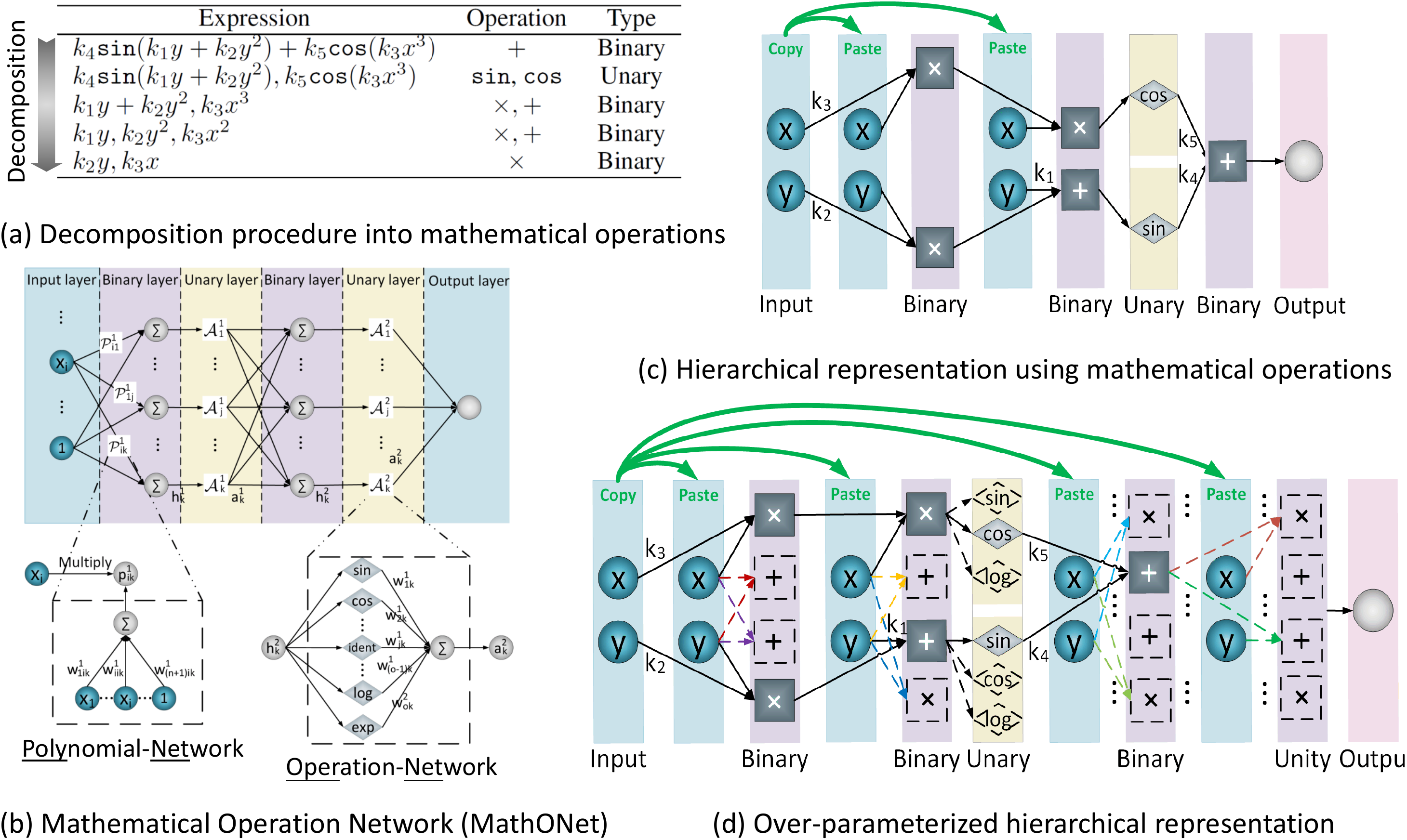}	
	\caption{ 
	  Mathematical Operation Network (MathONet).
		(a)  Decomposition procedure for $k_4 \texttt{sin}({k_1}y+{k_2}y^2)+{k_5}\texttt{cos}({k_3}x^3)$ into mathematical operations in term of unary operations (e.g., \texttt{sin},\texttt{cos},\texttt{log}, etc) and binary operations (e.g. $+,-,\times$, etc). 
		(b) A MathONet includes two basic modules, i.e., Polynomial-Network (PolyNet) and Operation-Network (OperNet). The stacked PolyNets form the binary layer, and the stacked OperNets form the unary layer. 
		(c) A DenseNet-like hierarchical representation using mathematical operations, where the square block stands for the binary operation and the diamond block represents the unary operation. The green line denotes that the input of the system will be copied and get involved in mathematical calculations performed by binary operations. 
		(d) An over-parameterized hierarchical structure. The black dotted connection represents the redundant connection that can be regularized on each weight. The dotted connections with different colors denote the redundant connections that can be regularized on a group of weights.
	}
	\vspace{-0.5cm}
	\label{fig:aritho_net}
\end{figure}

We will describe the motivation for MathONet design in more depth following the tutorial in Section~\ref{sec:intro} on the decomposition of $k_4 \texttt{sin}({k_1}y+{k_2}y^2)+{k_5}\texttt{cos}({k_3}x^3)$ into a DNN-like hierarchical representation (see Fig.~\ref{fig:aritho_net}(c) and (d)). As shown in Fig.~\ref{fig:aritho_net}(c), the hierarchical structure is stacked by two typical operations, i.e., unary operations (e.g. \texttt{sin},\texttt{cos},\texttt{log}, etc, denoted by diamond blocks) and binary operations (e.g. $+, \times$, etc, denoted by square blocks ). It should be noted that the input of the system will be ``copied'' (denoted as the green line in Fig.~\ref{fig:aritho_net}(c)) and ``pasted'' for the following layers (similar to the principle in DenseNet~\cite{huang2017densely}) for the calculations performed by binary operations. The design connects each layer to the preceding layers that can augment information flow and preserve the feed-forward nature similar to in DenseNet~\citep{huang2017densely}. The unary operations obtain input from preceding binary operations and pass their own feature-map to all subsequent binary operations.

As in Fig.~\ref{fig:aritho_net}(b), the MathONet is stacked by two typical layers, i.e., binary layer and unary layer. The binary layer is stacked by Polynomial-Network (PolyNet), and the unary layer is stacked by Operation-Network (OperNet). For a PolyNet, the system input is obtained as its input. Its output is a polynomial embodying the linear combination of input and is multiplied by the output of the previous layer. Each PolyNet can directly access the gradients from the loss function and system input by such design, thereby achieving implicit deep supervision. The unary layer is placed behind binary layer and performs linear (e.g., \texttt{ident}) or non-linear transformation(e.g., \texttt{sin, cos}) as its input. More detailed illustration for PolyNet and OperNet is in Sec.\ref{subsec:polyNet} and Sec.\ref{subsec:OperNet}, respectively. 

\vspace{-0.2cm}
\subsection{Polynomial-Network}
\label{subsec:polyNet}
\vspace{-0.2cm}
The Polynomial-Network (PolyNet) is a simple Fully-Connected network without hidden layers, as shown in Fig.~\ref{fig:aritho_net}(b).
Its input features are consist of two parts. The first part is the original input of the system. The second part is a constant which enables the MathONet to represent a linear system. Typically, the output of PolyNet is a polynomial embodying the linear combination of input as $p_{ik}^l=w_{(n+1)ik}^l+\sum\nolimits_{i=1}^n {(w_{iik}^l x_i)}$, 
where $\bx = [x_1, x_2, \ldots, x_n]\in R^{n}$ stands for the input of the system. $p_{ik}^l$ represents the output of the PolyNet $\mathcal{P}_{ik}^l$. $l$ is the layer index.
$w_{iik}^l$ is the weight parameter whose magnitude determines the strength of the connection within $\mathcal{P}_{ik}^l$. 
An example of PolyNet $\mathcal{P}_{ik}^1$ is illustrated in Fig.~\ref{fig:aritho_net}(b).

\vspace{-0.2cm}
\subsection{Operation-Network}
\label{subsec:OperNet}
\vspace{-0.2cm}
Operation-Network (OperNet) is designed as a linear combination of unary operations. As in Fig.~\ref{fig:aritho_net}(b), the general mathematical expression for the output of an OperNet is $a_k^l=\sum\nolimits_{o=1}^O f_o(w_{ok}^l h_k^l)$, 
where $h_k^l$ is the output of the $k_{th}$ neuron in layer $l$ with $h_k^l= \sum\nolimits_{i=1}^{n^l} (p_{ik}^l a_i^{l-1}+b_k^l)$. Specially for the first hidden layer, $h_k^1= \sum\nolimits_{i=1}^{n^1} (p_{ik}^1 x_i+b_k^1)$. 
$b_k^l$ is the bias. $f_o()$ stands for unary functions, e.g., ~\texttt{sin}, \texttt{cos}, \texttt{log}, \texttt{exp}.

By including the PolyNet and OperNet, the MathONet can be used to construct the governing equations of a dynamic system. 
The designed MathONet has the following advantages: a) a MathONet with even simple structure can represent a considerable expression space. For example, suppose an MathONet includes only $1$ hidden layer and $1$ hidden neuron with $2$ input features and $2$ unary functions, the compressed model can represent $2^{9}$ different expressions.
b) a MathONet can be trained with typical optimization method for neural networks, e.g. SGD.
c) a MathONet can approximate both linear and non-linear systems.
It should be noted that the complexity of a MathONet is mainly limited by three user-defined parameters: 
a) the number of hidden layers $L$, which denotes initialized model order and limits the depth of the MathONet;
b) $n^l$, the number of hidden neurons in layer $l$;
c) $O$, the number of unary functions for OperNet.

\section{Discovery Algorithm}
\label{sec:proposed algorithm}

This section will illustrate how to apply the Bayesian framework for a MathONet and explain the proposed sparse Bayesian deep learning algorithm. 
To ease notation, we use $\bw \in \mathbb{R}^{m \times 1}$ to represent a weight matrix within a PolyNet or OperNet.

\subsection{Sparse Group Lasso}
\label{subsec:spare_group_lasso}
Sparse group Lasso~\citep{simon2013sparse} is a regularization optimization technique which is a combination of Lasso~\cite{tibshirani1996regression} and group Lasso~\cite{yuan2006model}. Typically, suppose a weight matrix can be divided into different groups and $W_g$ represents a group of weights, its optimization target can be formulated as:
\begin{equation}
\label{eq:conventional loss function}
\min\limits_{{\bw}} {E}(\cdot) + \sum\nolimits_{\bw}\lambda \|{\bw}\|_{\ell_1} + \sum\nolimits_{\bw}\sum\nolimits_{g}\lambda_g \|{\bw_g}\|_{\ell_2}
\end{equation}
where $E(\cdot)$ is the energy function,
\begin{math} 
E(\cdot) = \frac{1}{\data}\sum\nolimits_{k=1}^{K} ({\by_{k} - \hat{\by_{k}}})^2 
\end{math}.  $\|{\bw}\|_{\ell_1}$ represents the Lasso regularization on each weight. And $\|{\bw_g}\|_{\ell_2}$ is the group Lasso regularization on a group of weights. In this paper, the weights within each PolyNet and OperNet can be collected as one group. Therefore the $\bw_g$ is equivalent to $\bw$ and $\|{\bw_g}\|_{\ell_2}$ can be represented as $\|{\bw_g}\|_{\ell_2} = \sqrt{\sum\nolimits_{i=1}^m(\bw_i^2)}$. 
$\lambda$ and $\lambda_g$ are the tuning parameters.

Before continuing, we would like to define two consistent terminologies (``epoch'' and ``cycle'') which will be used in the following Sections: one ``epoch'' refers to the entire dataset is processed forward and backward by the MathONet one time; each ``cycle'' includes $N_\text{epoch}$ epochs, and the network pruning is performed at the last epoch of each cycle.
As in~\citep{wen2016learning, kim2010tree}, sparse group Lasso can identify the redundancy of each connection and achieve structured sparsity by zeroing out all weights of a group. 
It should be noted in our algorithm, sparse group Lasso is exactly the first cycle to start. As shown in the Lorenz experiment, the identified model by sparse group Lasso is still redundant/non-sparse with $651$ terms (see Fig.~\ref{fig:lorenz_sparsity_loss_change_model_z}) and cannot reproduce attractor dynamics precisely (see Fig.~\ref{fig:lorenz_model_3_identified_arithonet_trajectory}(b)). 
Further research is needed to study why the sparse group Lasso is inefficient. These inefficiencies motivates us to develop a Bayesian learning version of sparse group Lasso, which can potentially yield sparser solutions.

\subsection{Sparse Group Bayesian Learning}
\label{subsec:bayesin_framework}
Given the dataset $\mathcal{D}=(\bx,\by) = {\{(\bx_k, \by_k)\}}_{k=1}^{K}$ and noise precision $\variance^{-2}$,
the likelihood for $\bw$ is assumed to be $
p(\by|\bw,\variance^2) = \prod_{k=1}^\data \mathcal{N}(\by_{k}| \text{Net}(\bx_k,\bw), \variance^2) 
$. A Gaussian prior distribution $P(W_i)$ is imposed on $W_i$ to regularize each weight. And a group prior $P(W_g)$ is imposed on $W_g$ to regularize a group of weights:
\begin{equation}
p(\bw) = \bN( \bw|\mathbf{0},\bG) =  \prod_{i=1}^{m}\bN(\bw_{i}|0,\hyper_{i})\\ \qquad
p(\bw_g) = \bN(\bw_{g}|\mathbf{0},\bG_{g}) = \prod_{i=1}^{m}\bN(\bw_{gi}|0,\hyper_{g})
\label{eq:gaussian prior}
\end{equation}

where \begin{math}
\bG = \diag\left[\Hyper \right]
\end{math},
$\Hyper \define \left[\hyper_1, \hyper_2, \ldots, \hyper_{m} \right] \in \R^{m \times 1}_{+}$ can be calculated by maximizing the model evidence~\citep{tipping2001sparse}, i.e., $ \Hyper = \argmax\nolimits_{\Hyper\geq\mathbf{0}} 
\int p({\by}|\bw,\sigma^2) \bN( \bw|\mathbf{0},\bG)d\bw
$ where
\begin{math}
\bG_g = \diag\left[\Hyper_g \right]
\end{math} and $\Hyper_g \define \left[\hyper_g, \hyper_g, \ldots, \hyper_{g} \right] \in \R^{m \times 1}_{+}$ is calculated by maximizing
$
\Hyper_g = \argmax\nolimits_{\Hyper_g\geq\mathbf{0}} 
\int p({\by}|\bw_g,\sigma^2) \bN( \bw|\mathbf{0},\bG_g)d\bw_g
$.
It should be noted that the parameter within $\Hyper_g$ shares the same value, 
which means that all weights of $\bw_g$ will be penalized identically.

To calculate the posterior distribution $p(\bw|\by)$, the model evidence $p(\by)$ is a necessity and can be evaluated by the integration $p(\by) = \int p(\by|\bw)p(\bw)d\bw$. However, the calculation of this integration is intractable and stands in need of approximation methods~\citep{tipping2001sparse,jacobs2018sparse}.
We consider the Laplace approximation approach.
$\Hyper$ and $\Hyper_g$ can be updated iteratively according to following iterative procedure:
\begin{align}
\beta = \sqrt{|\alpha|}, \quad \alpha = -{\mask}\times({1/{\Hyper}^2}) + 1/{{\Hyper}},
\quad \beta_g = \sqrt{|\alpha_g|}, \quad \alpha_g = \sum\nolimits_{i=1}^{m}(-{\mask_{gi}}/{\hyper_g^2} + 1/{{\hyper_g}})
\label{eq:cccp_1}
\end{align}
where 
\begin{math}
\mask = \left({\bG}^{-1}+{\Hessian}({\bw}, {\variance}^2)\right)^{-1} 
\end{math},
\begin{math}
\mask_{gi} = \left({\hyper_{g}^{-1}}+{\Hessian}({\bw_{gi}}, {\variance}^2)\right)^{-1} 
\end{math},
$\Hessian()$ denotes the Hessian matrix of $W$ or $W_{gi}$, respectively. 
$\Hyper$ and $\hyper_g$ can be calculated as $\Hyper = |\bw|/{\beta}$, $\hyper_g = \|{\bw_g}\|_{\ell_2}/{\beta_g}$. $\beta_g$ is a scalar and shared by all connections within the group $\bw_g$.

The common criteria of determining connection redundancy is based on the magnitude of weight, which is questionable as the magnitude does not indicate the optimal connection undoubtedly. 
In this work, we adopt $\alpha$ and $\alpha_g$ as the determining factor for connection redundancy.
As in~\eqref{eq:cccp_1}, the value of $\alpha$ and $\alpha_g$ is mainly decided by the uncertainty $\Hyper$, $\Hyper_g$ and $\Hessian$. 
Typically, the change of $\alpha$ ($\alpha_g$) is the opposite of $\Hyper$ ($\Hyper_g$).
An increase in $\Hyper$ ($\Hyper_g$) will cause $\alpha$ ($\alpha_g$) to decrease, thereby reducing regularization on corresponding weight $\bw$ ($\bw_g$). 
Based on this, the binary matrices ${C}$ and $C_g$ are generated as the \textit{masks} of $\bw$ and $\bw_g$, which denotes the connection redundancy and group redundancy, respectively.
$C$ ($C_g$) has the same dimension as $\bw$ ($\bw_g$) and will be optimized during the training process. The value is decided by:
\begin{equation}
\label{eq:mask update}
C = \left\{
\begin{array}{ll}
0, &  {\alpha}>\kappa_{\alpha}\\
1, &  \text{others} \\
\end{array}
\right.
\qquad \qquad
C_g = \left\{
\begin{array}{ll}
0, &  {\alpha_g}>\kappa_{\alpha_g}\\
1, &  \text{others} \\
\end{array}
\right.
\end{equation}
where $\kappa_{\alpha}, \kappa_{\alpha_g}$ stands for the thresholds for connection pruning and group pruning, respectively. $0$ denotes the redundancy, and $1$ means the weight should be retained. 
It should be noted that the \textit{masks} $C$ and $C_g$ will be updated at the last epoch of each cycle. 
The generic optimization target for the MathONet can be formulated as:
\begin{equation}
\label{eq:loss function}
\min\limits_{{\bw, C}} {E}(\cdot) + \sum\nolimits_{\bw,c}\lambda \|\beta \odot C \odot  {\bw}\|_{\ell_1} + \sum\nolimits_{\bw_g, C_g}\lambda_g \|\beta_g \cdot C_g \odot {\bw_g}\|_{\ell_2}.
\end{equation}
The dimension of ${C}$ and $\beta$ are the same as $\bw$. The dimension of $C_g$ is the same as $\bw_g$. $\beta_g$ a scalar which is shared by all connections within the group $\bw_g$.
$\odot$ denotes the Hadamard product. 
Inspired by \citep{gal2016uncertainty, loquercio2020general}, we also evaluate the predictive uncertainty using a practical Monte-Carlo sampling method. 
By sampling over the inferred posterior distribution for $T$ repetitions, 
an unbiased estimate of prediction can be approximated by the average of predicted output, 
i.e., \begin{math}
\bar{\by_{k}} =  \frac{1}{T}\sum\nolimits_{t=1}^{T} \hat{\by_k^t}
\end{math}. 
where $\hat{\by_k^t}$ is the predicted output of the $t$-th samples and \begin{math}
\Sigma_{\hat{\by_k}} = \frac{1}{T}\sum\nolimits_{t=1}^{T} (\hat{\by_k^t} - \bar{\by_k})^2
\end{math},
Suppose $N_{\text{cycle}}$ represents the maximal cycles and $N_{\text{epoch}}$ denotes the number of epochs in each cycle, a pseudo-code for the discovery algorithm is given by Algorithm~\ref{algo:algorithm}. The detailed proof for~\eqref{eq:cccp_1} and~\eqref{eq:mask update} is in Appendix~\ref{appsec:cccp_procedure derive}.

\begin{algorithm}[ht] 
	\caption{Bayesian learning discovery algorithm}
	\label{algo:algorithm}
	\textbf{Initialize:} hyper-parameters $\beta, \hyper = I$
	;  regularization tuning parameter $\lambda \in \R^{+}$; threshold for pruning $\kappa_\alpha,\kappa_{\alpha_g} \in \R^{+}$. $N_\text{cycle} \in Z^{+}$ denotes the maximum cycles; $N_\text{epoch}\in Z^{+}$ denotes the number of epochs in each cycle.
	
	\textbf{for} {$i = 1$ to $N_{\text{cycle}}$}

	 \qquad \textbf{for} {$j = 1$ to $N_\text{epoch}$}
	 
	 	\qquad \qquad 1. Update the weight $\bw$ by minimizing loss function as ~\eqref{eq:loss function}.

	 \qquad \textbf{end for} 
	 
	\begin{enumerate} \setcounter{enumi}{1}
		\item Update hyper-parameters $\Hyper$ as~\eqref{eq:cccp_1}.
		\item Update mask $C$ and $C_g$ as~\eqref{eq:mask update}.			
	\end{enumerate} 
	\textbf{end for} 
\end{algorithm}

\section{Experimental Result}
\label{sec:result}
In this section, the algorithm is demonstrated on the chaotic Lorenz system~\cite{lorenz1963deterministic}, Lotka-Volterra system~\cite{berryman1992orgins} and Kolmogorov–Petrovsky–Piskunov (Fisher-KPP) system~\cite{fisher1937wave}. 

\subsection{Chaotic Lorenz System}
\label{sebsec:Lorenz System}
\begin{figure}[ht]
	\centering
	\includegraphics[trim=19.5cm 1cm 10.5cm 1cm,scale = 0.14]{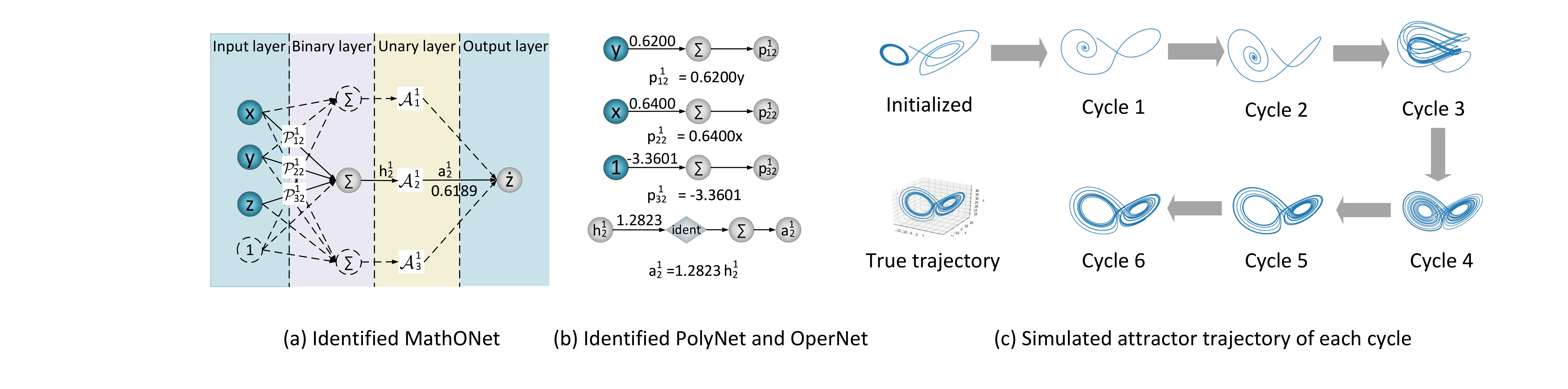}
	\caption{
		Identified result of Lorenz system. (a) Identified MathONet model for model $z$. The MathONet is initialized with $1$ hidden layer and $3$ hidden neurons. 
		The OperNet includes $5$ basic unary functions in the beginning, i.e.,~\texttt{identity}, \texttt{sin}, \texttt{cos}, \texttt{log}, \texttt{exp}. Algorithm~\ref{algo:algorithm} can identify the essential connections after $6$ cycles. The dotted parts are the identified redundant connections and neurons. (b) The identified PolyNets and OperNets represent simple mathematical expressions that are placed under each basic modules. 
		(c) The attractor trajectories generated by the intermediate model of each cycle. The identified model can reproduce the attractor dynamics as the true model after $6$ cycles. 
	}
	\label{fig:lorenz_model_3_identified_arithonet_trajectory}
\end{figure}

As a typical canonical model for chaotic dynamics, the Lorenz system is non-linear, non-periodic and is notable for its chaotic solution being sensitive to system parameters and initial conditions. Although the dynamics of the Lorenz attractor is difficult to interpret, the attractor action can be described by a simplified mathematical model, which is a three-dimensional and deterministic ODE: 
\begin{equation}
	\Dot{x} = \sigma(y - x),
	\qquad
	\Dot{y} = x(\rho - z) - y,\qquad
	\Dot{z} = xy - \beta z
	\label{eq:theoretical lorenz}
\end{equation}
With the standard parameter values ${\sigma =10}, {\beta =8/3}$ and ${\rho =28}$, the system exhibits chaotic behavior as shown in Fig.~\ref{fig:lorenz_model_3_identified_arithonet_trajectory}(c). The data is collected through simulation experiments. The state vector and their derivatives are stacked as input and output dataset, respectively. 

Due to space limitation, we only show the identified result for model $z$ in Fig.~\ref{fig:lorenz_model_3_identified_arithonet_trajectory}. 
Fig.~\ref{fig:lorenz_model_3_identified_arithonet_trajectory}(b) shows the trajectory of the attractor represented by the model generated in each cycle during the training process.
The first cycle (the definition of cycle is in Algorithm~\ref{algo:algorithm}) is the conventional regularization that cannot identify the dynamics precisely. By applying the Bayesian learning algorithm~\ref{algo:algorithm}, the identified model of cycle $6$ can reproduce the attractor dynamics accurately.

\begin{figure}
	\centering
	\vspace{-0.5cm}
	\begin{subfigure}[b]{0.44\textwidth}
		\centering
		\includegraphics[height = 3.5cm]{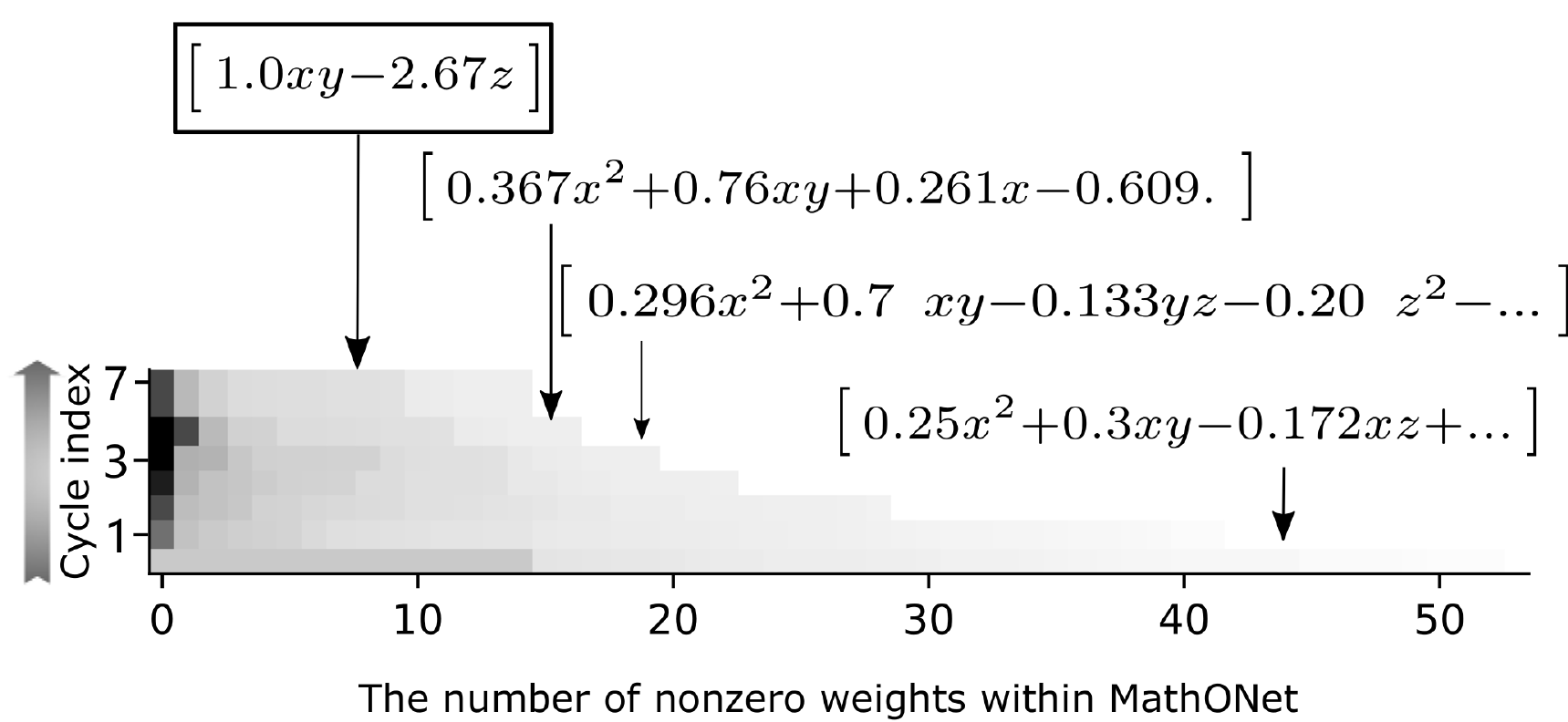}
		\caption{The number of nonzero weights of the MathONet generated in each cycle.}
		\label{fig:lorenz_weight_equation_change_model_z}
	\end{subfigure}
	\hspace{0.1\textwidth}
	\begin{subfigure}[b]{0.44\textwidth}
		\centering
		\includegraphics[height = 3.8cm]{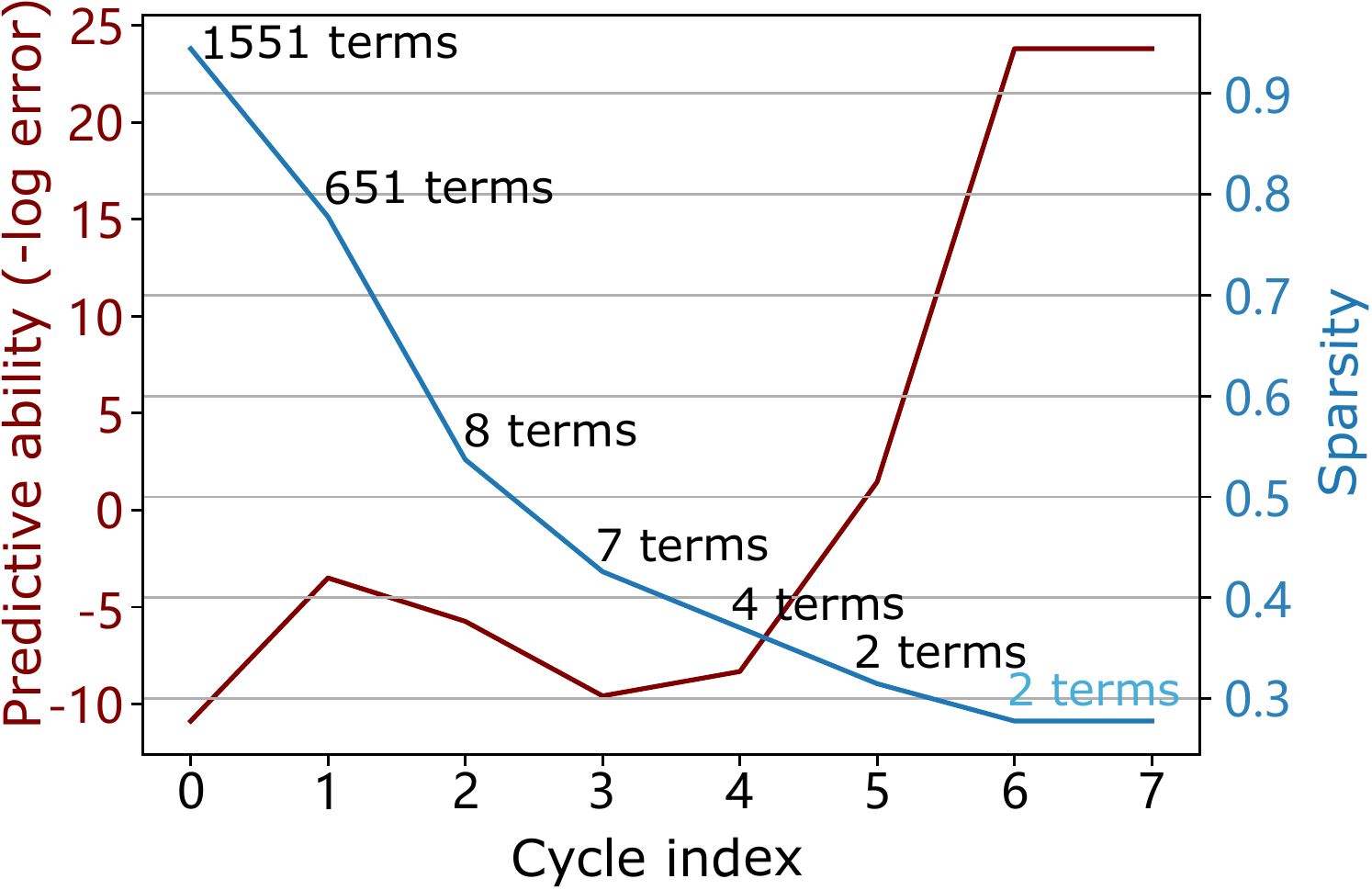}		
		\caption{The sparsity and predictive ability of the MathONet generated in each cycle.}
		\label{fig:lorenz_sparsity_loss_change_model_z}
	\end{subfigure}   
	\caption{The sparsity, predictive ability and weights of the identified MathONet generated in each cycle, which aims to discover governing equation of model $z$ of Lorenz system.
		a) The nonzero weights of the MathONet generated in each cycle. 
        The horizontal axis represents the combination of non-zero weights in the MathONet generated in each cycle.
		The vertical axis denotes the index of training cycles.
	    The expression at each turning line (the cliff) represents the governing equation identified in the corresponding cycle.
		b) The sparsity and prediction ability of the MathONet identified in each cycle. The model becomes more and more sparse and has more and more predictive ability. The annotation next to the sparsity line represents the number of identified mathematical terms of the corresponding cycle. The first cycle represents the result identified by sparse group Lasso method which is still redundant ($651$ terms), and the prediction ability is low.
	}
	\label{fig:lorenz_weight_sparsity_loss_change_model_z}
\end{figure}

Fig.~\ref{fig:lorenz_weight_sparsity_loss_change_model_z} shows the process of the discovering governing equations for model $z$. 
In Fig.~\ref{fig:lorenz_weight_sparsity_loss_change_model_z}(a), only the non-zero weight elements within the MathONet are collected to form a weight vector in each cycle. The identified governing equation has undergone a transformation from complex to simple. 
If combined with Fig.~\ref{fig:lorenz_weight_sparsity_loss_change_model_z}(b), which shows the change of predictive ability and model sparsity, it can be observed that the predictive ability tends to improve as the model complexity decreases.
The initialized structure has $1551$ terms in total, while only $2$ terms are retained after $6$ cycles.
This also verifies that the proposed method can discover governing equations in a reasonable, effective and accurate manner.
To explore the robustness of the proposed method with noisy derivatives measurement, a Gaussian noise $\noise = \prob(0,\variance^2)$ with $\variance \in \{0.01, 1, 10\}$ is added to the exact derivatives, respectively. The result shows that linear and quadratic terms are correctly identified even under the large noise value ($\variance = 10$). 
Besides, although the simulation accuracy of attractor dynamics decreases with the increase of noise, the coefficients $\sigma, \beta, \gamma$ can still be determined accurately within $0.5\%$ around the true value. The detailed identified structure and parameters can be found in Table~\ref{tab:lorenz_noisy_result} in Appendix~\ref{subsec:app_result_lorenz}. The prediction uncertainty for noisy datasets can be found in Fig.~\ref{fig:lorenz_uncertainty}.

\subsection{Lotka-Volterra System}
\label{subsubsec:Lotka-Volterra}
The Lotka-Volterra system is known as a general framework of an ecological system that can describe the dynamic relationship between the natural population of a predator and prey through time~\cite{berryman1992orgins}. With two main assumptions: a) the growth rate of population is proportional to its size;  b) The predator population can only get food from the prey population, and the food for the prey population is supplied sufficiently at all times. The Lotka-Volterra equation can be described by a pair of deterministic first-order non-linear ODEs:
\begin{equation}
	\Dot{x} = \alpha x - \beta xy,
	\qquad
	\Dot{y} = \delta xy - \gamma y
	\label{eq:theoretical Lotka Volterra}
\end{equation}where $\dot{x}$ and $\dot{y}$ represents the instantaneous growth rates of the prey ($x$) and predator ($y$), respectively. 
It can be found that the growth rate of each species is determined by two factors, i.e., the population of itself and the interaction with the other species. 
\begin{figure}[ht]
	\centering
	\includegraphics[trim=10.5cm 6cm 10.5cm 3cm,scale = 0.18]{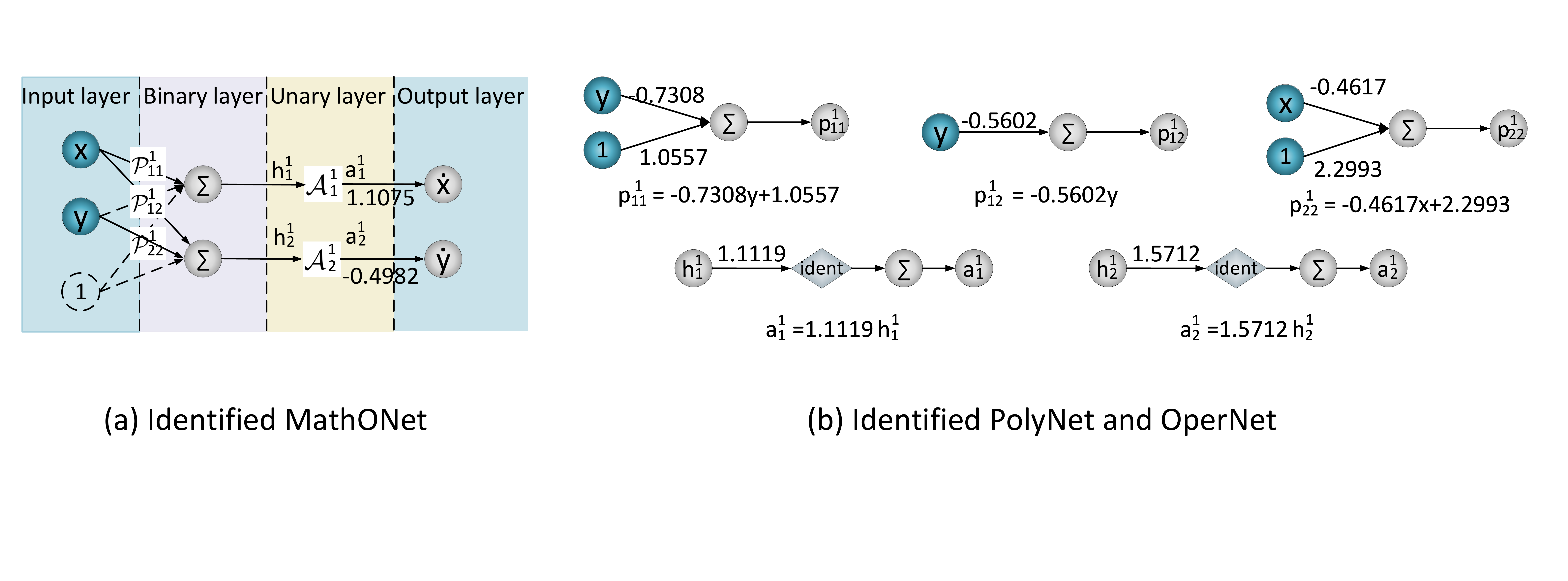}
	\caption{
		Identified MathONet model for Lotka-Volterra system. The MathONet is initialized with $1$ hidden layer and $2$ hidden neurons. 
		The OperNet includes $3$ unary functions, i.e.~\texttt{identity},\texttt{sin},\texttt{cos}. 
		(b)The identified PolyNets and OperNets, where the dotted line are the identified redundant parts. 
	}
	\label{fig:lk_identified_arithonet}
\end{figure}
Fig.~\ref{fig:lk_identified_arithonet} shows the identified MathONet. It can be found that the introduced constant input is removed from the input layer, which is in accordance with~\eqref{eq:theoretical Lotka Volterra}. The PolyNet between input feature $y$ and the first hidden neuron is also identified to be redundant. Although the other basic modules are retained, the identified structures are sparse and equivalent to simple mathematical expression, as shown in Fig.~\ref{fig:lk_identified_arithonet}(b). We also investigate the simulated output and predicted distribution of the identified model, which is shown in Fig.~\ref{fig:lk_solution} of Appendix.Sec.~\ref{appsubsubsec:Lotka-Volterra_detials}.
The change of identified governing equations during the searching process, the change of sparsity and predicted ability are in Fig.~\ref{fig:Lotka_Volterra_weight_sparsity_loss_change} in Sec.~\ref{appsubsubsec:Lotka-Volterra_result} of the Appendix.

\subsection{Fisher-KPP (Kolmogorov–Petrovsky–Piskunov) Equation}
\label{subsubsec:fisher}
Fisher equation, also known as Kolmogorov–Petrovsky–Piskunov (KPP) equation, is a typical semilinear reaction-diffusion equation that describes the population growth and propagation of a species~\cite{fisher1937wave}. The spatio-temporal dynamics of Fisher-KPP can be represented by a PDE: 
\begin{equation}
\begin{aligned}
\label{eq:theoretical fisher}
\frac{\partial{p}}{\partial{t}} = 
d\frac{\partial^2{p}}{\partial{x^2}}+rp(1-p)
\end{aligned}
\end{equation}
where $p$ denotes the population density. $x \in [0,1]$ represents the coordinate measurement position of a species. $t \in [0,T]$ stands for the time in generations. $d = 6.25$ is a constant denoting the diffusion coefficient. $r = 1$ represents the intensity in favor of local population growth, assumed to be independent of $p$.
The dataset is generated by the Julia code provided in~\cite{rackauckas2020universal}. The dimension of the dataset is $26 \times 11$, in which $11$ refers to the number of samples for time; and $26$ refers to the number of samples for the position. 

\begin{figure}[ht]
	\centering
	\includegraphics[trim=10.5cm 28cm 17.5cm 2cm,scale = 0.15]{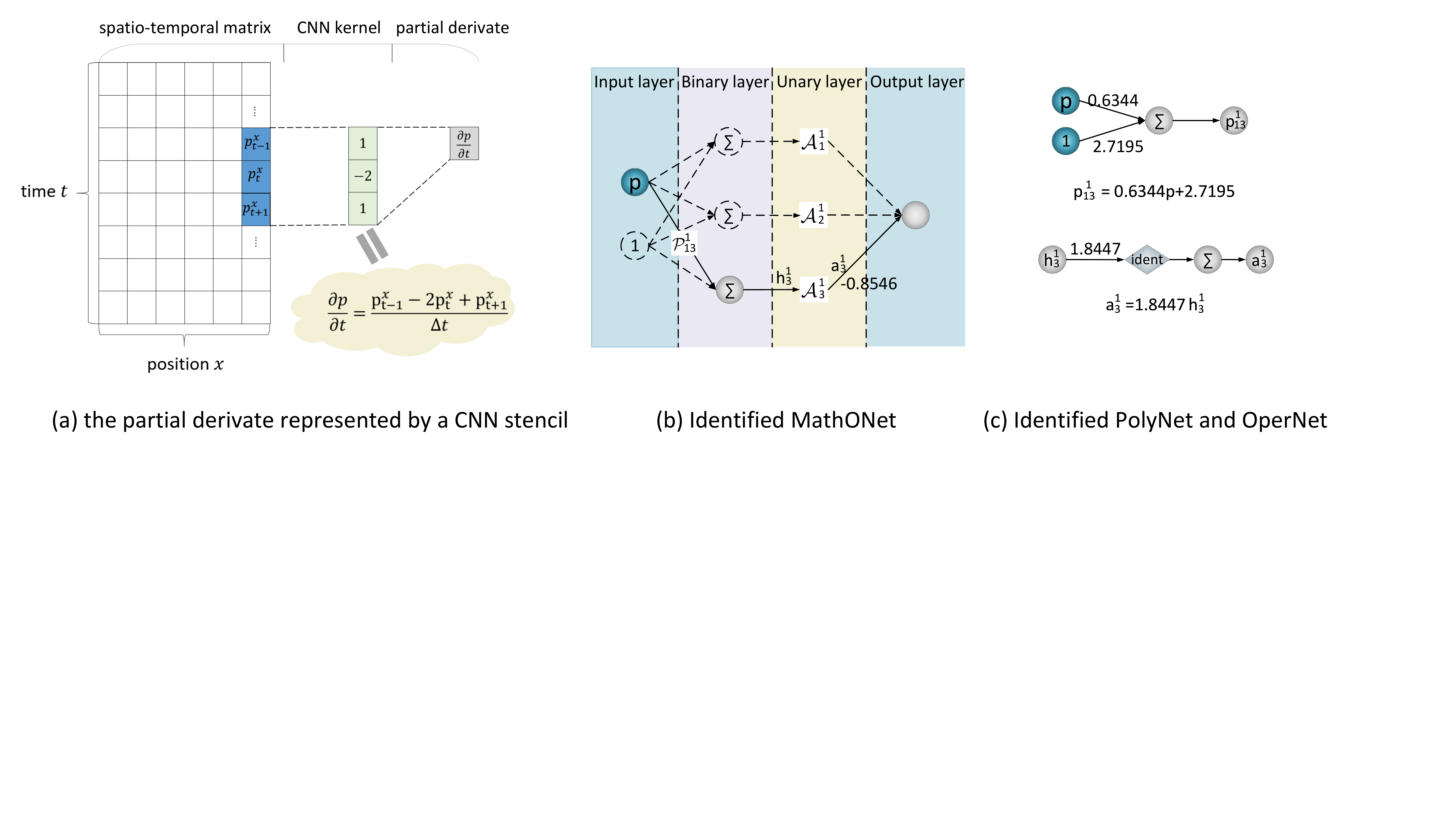}
	\caption{
		An illustration to model the PDE by combining a MathONet and a CNN with a special kernel. 
		(a) An explanation of the partial derivative $\frac{\partial{p}}{\partial{t}}$ represented by a CNN stencil with kernel $[1,-2,1]$.
		(b) Identified MathONet model for Fisher-KPP system. The MathONet is initialized with $1$ hidden layer and $3$ hidden neurons. The OperNet includes $3$ basic unary functions, i.e.~\texttt{identity},\texttt{sin},\texttt{cos}. Only two basic modules, i.e., $\mathcal{P}_{13}^1$ and $\mathcal{A}_{3}^1$ are retained in the graph. (c) Identified PolyNet and OperNet.
	}
	\label{fig:fisher_identified_arithonet}
\end{figure}

As shown in \eqref{eq:theoretical fisher}, the one-dimensional Fisher equation consists of two parts, i.e., a derivative operator and a polynomial representing the local growth item. In this experiment, we use a MathONet to approximate the local growth item. 
Inspired by~\cite{rackauckas2020universal, ruthotto2019deep}, a discretized PDE can be interpreted as a convolutional layer that can exploit the relation between adjacent elements of a 2-D matrix~\citep{raina2009large}. In Fig.~\ref{fig:fisher_identified_arithonet}(a), we explain this affine transformation on a spatial-temporal matrix. It should be noted that an ideal CNN stencil should be $[1,-2,1]$, which satisfies the physical constraint that the sum of CNN stencil being zero.
To ensure the physical interpretability of the learned CNN stencil, previous work~\cite{rackauckas2020universal} imposed the physical constraint on the CNN kernel and obtained the desired result. 
In this work, we aim to discover the governing equations only from data. Therefore, we only include a simple convolutional neural network with the kernel size $3 \times 1$ is and try to learn the stencil.

The optimal learned model is obtained with the identified equation:
$$\frac{\partial{p}}{\partial{t}} = 
3.2424*\text{CNN}(p)-1.5765p(0.6344*p + 2.7195)$$
The identified CNN stencil is $[1.9276,-2.2245,1.9276]$. If this stencil is re-scaled to $[1.000, -2.000, 1.000]$, the equivalent equation will be
\begin{math}
\frac{\partial{p}}{\partial{t}} = 
6.250*\text{CNN}(p)+1.000p(1.000-1.000p)
\end{math} which is almost the same as the true governing equations. 
Therefore, both the structure and coefficients of~\eqref{eq:theoretical fisher} can be accurately learned. The identified structure of MathONet is in Fig.~\ref{fig:fisher_identified_arithonet}(b).
We also investigate the change of identified governing equations, model sparsity, predicted ability along the training process, and the prediction uncertainty of the identified model. A more detailed result is shown in Fig.~\ref{fig:fisher_weight_sparsity_loss_change} of Appendix.Sec.~\ref{subsec:app_result_fisher}.

\section{Conclusion}
\label{sec:conclusion}

We present a method that can learn the mathematical operations in governing equations of dynamic systems composed of the basic mathematical operations, i.e., unary and binary operations.
The governing equations are formulated as a DenseNet-like hierarchical structure, termed as MathONet. The governing equations discovery problem can be formulated as a deep neural network compression problem with redundant mathematical operations. A Bayesian learning approach is proposed to find a sparser solution compared with sparse group Lasso-type algorithms. The experiment result shows that the proposed method effectively discovers general differential equations, including linear and non-linear differential equations, ordinary differential equations (ODEs), or partial differential equations (PDEs).

\bibliographystyle{unsrt}

\newpage
\newpage
\appendix
\section{Sparse Group Bayesian Learning Algorithm Derivation}
\label{appsec:cccp_procedure derive}
As explained in Sec.~\ref{subsec:bayesin_framework}, the Laplace approximation method is adopted to calculate the intractable integral for the model evidence \begin{math}
p(\by) = \int p(\by|\bw)p(\bw)d\bw
\end{math}, where $\bw \in \mathbb{R}^{m \times 1}$ represents a weight matrix within a PolyNet or OperNet.

The sparse group Bayesian learning algorithm is a Bayesian learning version of the sparse group Lasso. It imposes the Gaussian prior distribution on each weight as the non-structured regularization and imposes group prior on a group of weights as the structured regularization.
In this section, a more detailed mathematical description of the sparse group Bayesian learning algorithm is given.

\subsection{Non-structured regularization}
\label{subappsec:non-structured regularization}
\subsubsection{Laplace Approximation}
\label{appsubsec:laplace_approximation_non-structured_regularization}
Given the dataset $\mathcal{D}=(\bx,\by) = {\{(\bx_k, \by_k)\}}_{k=1}^{K}$ and noise precision $\variance^{-2}$, the likelihood is defined by a Gaussian distribution: 
\begin{equation}
p(\by|{\bw},\sigma^2) = \prod_{k=1}^\data \mathcal{N}(\by_{k}| \text{Net}(\bx_k,\bw) =   (2\pi\sigma^2)^{\frac{\data}{2}}\exp\left(-\frac{1}{2\sigma^2}{\sum_{k=1}^{\data}\left(Y_{k}-\text{Net}\left(\bx_k,{{\bw}}\right)\right)^2}\right)
\label{eq:general:generallikelihood}
\end{equation}
Based on Eq.~\eqref{eq:gaussian prior}, the prior is 
\begin{equation}
p(\bw) = \bN( \bw|\mathbf{0},\bG) =  \prod_{i=1}^{m}\bN(\bw_{i}|0,\hyper_{i})
\label{eq:general:generalprior}
\end{equation}

To compute the intractable integral for the evidence as in~\eqref{general:generalmarignal}, the energy function  \begin{math}E({\bw},\sigma^2)=\energy \end{math} is expanded according to a second-order Taylor series expansion around $\bw^*$:
\begin{equation}
\begin{aligned}
E({\bw},\sigma^2)
\approx E(\bw^*, \btheta)+({\bw}-\bw^*)^{\top}\mathbf{g}(\bw^*, \btheta)+\frac{1}{2}({\bw}-\bw^*)^{\top} \mathbf{H}(\bw^*, \btheta) ({\bw}-\bw^*) 
\label{eq:taylor_energy_function}
\end{aligned}
\end{equation}
where $\mathbf{g}(\cdot)=\nabla \mathbf{E}(\bw,\sigma^2)|_{\bw^*}$ is the gradient and $\mathbf{H}(\cdot) = \nabla\nabla \mathbf{E}(\bw,\sigma^2)|_{\bw^*}$ is the Hessian.
With this expansion, the likelihood function~\ref{eq:general:generallikelihood} can be rewritten as:
{\small{
	\begin{equation}
	\begin{aligned}
	& p({\by}|{\bw}, \btheta) 
	= {(2\pi\sigma^2)^{\frac{\data}{2}}} \cdot \exp\{-\energy\} \\
	\approx & {(2\pi\sigma^2)^{\frac{\data}{2}}} \cdot \exp \left \{-\left(\quadratic
	+E(\bw^*, \btheta)\right)\right \} \\
	= & a(\bw^*, \btheta) \cdot \exp \left \{-\left(\frac{1}{2}\bw^{\top} \Hessian(\bw^*, \btheta) {\bw}+ \bw^{\top} \hat{\Gradient}(\bw^*, \btheta)\right) \right \} 
	\label{eq:likelihood-approximation}
	\end{aligned}
	\end{equation}}}
with
\begin{equation}
\begin{aligned}
\hat{\Gradient}({\bw}^*, \btheta)& \define \Gradient(\bw^*, \btheta)-\Hessian(\bw^*, \btheta) \bw^*.
\notag
\\
a(\bw^*, \btheta) & \define {(2\pi\sigma^2)^{\frac{\data}{2}}}  \exp \left \{- \left(\frac{1}{2}{\bw^*}^{\top} \Hessian(\bw^*, \btheta) {\bw^*} - {\bw^*}^{\top} \Gradient(\bw^*, \btheta)  + E(\bw^*, \btheta) \right) \right \} 
\end{aligned}
\end{equation}

Given the Gaussian prior~\eqref{eq:general:generalprior} and the approximated likelihood~\eqref{eq:likelihood-approximation}, the posterior is also a Gaussian $\mathcal{N}(\mean_{{\bw}},{\Sigma}_{{\bw}}$ by the effect of the conjugacy rule:
\begin{equation}
	\mean_{{\bw}} = {{\Sigma}_{{\bw}}} \cdot \left[\Gradient(\bw^*, \btheta) + \Hessian(\bw^*, \btheta) \bw^* \right], \quad {{\Sigma}_{{\bw}}}= \left[\Hessian(\bw^*, \btheta)+ \bG^{-1} \right]^{-1}
\end{equation}

\subsubsection{Evidence Maximization}
\label{appsubsec:evidence_maximization}
As in David MacKay's Bayesian framework~\cite{mackay1992bayesian}, the model evidence can be approximated by the posterior volume, i.e. 
\begin{math}
	p(\by|{\bw},\sigma^2)p({\bw}) \approx p(\by|{\mean_{{\bw}}},{\Sigma}_{{\bw}})p({\mean_{{\bw}}})
\end{math}.
Based on~\eqref{eq:general:generalprior} ~\eqref{eq:likelihood-approximation}, the model evidence is:
\begin{equation}
\begin{aligned}
&p(\by) = \int p(\by|{\bw},\sigma^2)p({\bw})d{\bw}
=  \int p({\by}|{\bw}, \btheta) \bN( {\bw}|\mathbf{0},{\bG})d{\bw} \\
=&\frac{a(\bw^*, \btheta)}{\left(2\pi\right)^{m/2}|{\bG}|^{1/2}}
\int \exp\{\frac{1}{2}\bw^{\top} \Hessian(\bw^*, \btheta) {\bw}+ \bw^{\top} \hat{\Gradient}({\bw}^*, \btheta)+\frac{1}{2} \bw^{\top} \bG^{-1} {\bw}\}d{\bw} \\
\propto & \frac{a(\bw^*, \btheta)}{\left(2\pi\right)^{m/2}|{\bG}|^{1/2}}
\exp\{\frac{1}{2}\bw^{\top} \Hessian(\bw^*, \btheta) {\bw}+ \bw^{\top} \hat{\Gradient}({\bw}^*, \btheta)+\frac{1}{2} \bw^{\top} \bG^{-1} {\bw}\} |\Sigma_\bw|^{\frac{1}{2}} d{\bw} 
\label{general:generalmarignal}
\end{aligned}
\end{equation}

Applying a $-2\log(\cdot)$ transformation to~\eqref{general:generalmarignal}: 
{\small{
\begin{equation}
\begin{aligned}
&-2\log\left[\frac{a(\bw^*, \btheta)}{\left(2\pi\right)^{m/2}|{\bG}|^{1/2}}
\exp\{\frac{1}{2}\bw^{\top} \Hessian(\bw^*, \btheta) {\bw}+ \bw^{\top} \hat{\Gradient}({\bw}^*, \btheta)+\frac{1}{2} \bw^{\top} \bG^{-1} {\bw}\} |\Sigma_\bw|^{\frac{1}{2}} d{\bw} \right] \\
\propto & -2 a(\bw^*, \btheta) +\log  |{\bG}|+\log |\Hessian(\bw^*, \btheta)+ \bG^{-1}| + \frac{1}{2}\bw^{\top} \Hessian(\bw^*, \btheta) {\bw}+ \bw^{\top} \hat{\Gradient}({\bw}^*, \btheta) \\
&+\frac{1}{2} \bw^{\top} \bG^{-1} {\bw} 
\\
\propto & \bw^{\top} \Hessian(\bw^*, \btheta) {\bw}+ 2\bw^{\top} \hat{\Gradient}(\bw^*, \btheta)+ \bw^{\top} \bG^{-1} {\bw} +\log  |{\bG}|+\log |\Hessian(\bw^*, \btheta)+ \bG^{-1} |.
\label{general:prior:integral3}
\end{aligned}
\end{equation}}
}
Therefore, the maximization of evidence becomes minimizing following objective function:
\begin{equation}
\begin{split}
&\mathcal{L}(\bw, \bgamma) = \bw^{\top} \Hessian \bw+ 2 \bw^{\top}(\Gradient-\Hessian \bw^{*})
+ \bw^{\top} \bG^{-1} \bw + \log |\bG| + \log |\Hessian+ \bG^{-1} |	
\end{split}
\end{equation}

\subsubsection{Regularization Update Rules}
\label{appsubsec:Regularization Update Rules}
\begin{prop}
	\label{general:proposition2:cost}
	With known $\variance$,the $\bw$ and $\hyper$ can be optimized by minimizing the objective function:
	\begin{equation}
	\label{eq:general:cost:cccp}
	\begin{split}
	&\mathcal{L}(\bw, \bgamma) = \bw^{\top} \Hessian \bw+ 2 \bw^{\top}(\Gradient-\Hessian \bw^{*})
	+ \bw^{\top} \bG^{-1} \bw + \log |\bG| + \log |\Hessian+ \bG^{-1} |	
	\end{split}
	\end{equation}
\end{prop}
where \begin{math}
u(\bw,\bG) = \bw^{\top} \Hessian \bw+ 2 \bw^{\top}(\Gradient-\Hessian \bw^{*})
+ \bw^{\top} \bG^{-1} \bw
\end{math} is convex jointly in $\bw,\bG$. And \begin{math}
v(\bG) = \log |\bG| + \log |\Hessian+ \bG^{-1} | 
\end{math} is concave in $\bG$. The optimization problem can be solved with a convex-concave procedure (CCCP).

\begin{pf}
	Eq~\eqref{eq:general:cost:cccp} is consist of a convex part in $\bw, \bG$ and a concave part in $\bG$. For the convex part:
		\begin{equation}
		\begin{aligned}
		{u}\left(\bw, \bG \right) 
		= \bw^{\top} \Hessian \bw+ 2 \bw^{\top}(\Gradient-\Hessian \bw^{*}) + \bw^{\top} \bG^{-1} \bw
		\label{general:summary:function:u}
		\end{aligned}
		\end{equation}
	${u}\left(\bw, \bG \right)$ is a convex function as it is the sum of convex functions with type $f(\mathbf{x}, Y) = \mathbf{x}^{\top} \mathbf{Y}^{-1} \mathbf{x}$ \cite{boyd2004convex}.
	By using the Schur complement determinant identities, the concave part can be written as a $\log$-determinant of an affine function of semidefinite matrices $\bG$: 
	\begin{equation}
	\begin{aligned}
	v(\bG) & =  \log |\bG| + \log|{\bG}^{-1}  + {\Hessian}(\bw^*, \btheta)|   =   \log \left(|\bG|| {\bG}^{-1}  + {\Hessian}(\bw^*, \btheta)| \right)  \\
	& = \log \left| \begin{pmatrix}
	{\Hessian}(\bw^*, \btheta) & \\
	& -\bG
	\end{pmatrix} \right|  = \log\left|\bG +  {\Hessian}^{-1}(\bw^*, \btheta) \right| + \log \left|{\Hessian}(\bw^*, \btheta) \right|
	\label{eq:matrix determinant expansion}
	\end{aligned}
	\end{equation}
	
	Therefore, the optimization problem in Eq.~\eqref{eq:general:cost:cccp} could be solved with a convex-concave procedure (CCCP). Specifically, by computing the gradient of $v(\bgamma)$ to $\bG$ and the gradient of $u(\bw,\bgamma)$ to $\bw$, we have the following iterative optimization procedure:	
		{
		\begin{equation}
		\bgamma =\argmin\limits_{\bgamma \succeq \mathbf{0}} u(\bw, \bgamma)+\nabla_{\bgamma} v(\bgamma)^\top\bgamma \label{eq:general:summary:eq:cccp2}
		\end{equation}
	}
	{
		\begin{equation}
		\label{eq:general:summary:eq:cccp1}
		\bw =\argmin\limits_{\bw} u(\bw, \bgamma)
		\end{equation}
	}	
\end{pf}
Based on \eqref{eq:general:summary:eq:cccp2}, we could update ${{\bG}}$. 
By using the chain rule, its analytic form is:
{
	\begin{equation}
	\begin{aligned}
	\balpha
	\define &\nabla_{{\bG}} v\left({\bG}\right)^\top  |_{{\bG}={\bG^*}}=\nabla_{{\bG}}\left(\log |\bG^{-1}+{\Hessian}(\bw^*, \btheta) |+\log  |{\bG}| \right)^\top|_{{\bG}={\bG^*}}\\
	=& -\diag \left \{\bG^{-1} \right \} \circ  \diag \left \{\left(\bG^{-1}+{\Hessian}(\bw^*, \btheta)\right)^{-1} \right \} \circ \diag \left \{\bG^{-1} \right \}  +\diag \left \{\bG^{-1} \right \} \\
	=& 
	\diag\left \{\left[
	\begin{array}{ccc}
	\alpha_{1}& \cdots &  \alpha_{m}
	\end{array}
	\right]\right \}
	\label{eq:general:alpha1k}
	\end{aligned}
	\end{equation}
}where $\circ$ represents the Hadamard product; 
$\balpha$ is an intermediate variable. According to \eqref{eq:general:alpha1k}, each $\alpha_{i}$ is calculated as:
\begin{align}
\mask = \left(\bG^{-1}+{\Hessian}(\bw^*, \btheta)\right)^{-1} 
\qquad
{\alpha} = { -\frac{\mask}{(\bG)^2} +\frac{1}{\bG}} 
\label{eq:general:summary:gammastarupdate2}
\end{align}
Therefore, ${\bG}$ can be updated as ~\eqref{eq:general:summary:eq:cccp2}:
{
	\begin{equation}
	\begin{aligned}
	&\bgamma =\argmin\limits_{\bgamma \succeq \mathbf{0}} u(\bw, \bgamma)+\nabla_{\bgamma} v(\bgamma)^\top\bgamma \\
	= & \argmin\limits_{\bgamma \succeq \mathbf{0}} \bw^{\top} \Hessian \bw+ 2 \bw^{\top}(\Gradient-\Hessian \bw^{*}) + \bw^{\top} \bG^{-1} \bw + \nabla_{\bgamma} v(\bgamma)^\top\bgamma \\
	= & \argmin\limits_{\bgamma \succeq \mathbf{0}} \bw^{\top} \bG^{-1} \bw + \nabla_{\bgamma} v(\bgamma)^\top\bgamma \\
	\label{general:summary:cccp-4}
	\end{aligned}
	\end{equation}
}
Since $\frac{\bw_{i}^2}{\br_i} +\alpha_i\br_i \geq 2\left|\sqrt{\alpha_i} \cdot \bw_{i}\right|,$ the optimal ${\br_i}$ equals:
\begin{equation}
\begin{aligned}
\br_i=\frac{|\bw_{i}|}{\sqrt{\alpha_i}},  \forall i.
\label{eq:general:summary:cccp-5}
\end{aligned}
\end{equation}
Define \begin{math}
\beta \define \sqrt{\alpha}
\end{math},
${\bw}$ can be solved according to Eq~\eqref{eq:general:summary:eq:cccp1}:
\begin{equation}
\begin{aligned}
{\bw} =&\argmin\limits_{{\bw}}
\frac{1}{2}\bw^{\top} \Hessian {\bw}+ \bw^{\top}(\Gradient-\Hessian \bw^{*})  +\sum_{i=1}^{m} \|\beta \cdot \bw_{i}\|_{\ell_1} \\
\propto&\argmin\limits_{\bw} E(\bw^*, \btheta)+({\bw}-\bw^*)^{\top}\mathbf{g}(\bw^*, \btheta)+\frac{1}{2}({\bw}-\bw^*)^{\top} \mathbf{H}(\bw^*, \btheta) ({\bw}-\bw^*)  
\\
&+ 2\sum_{i=1}^{m} \|\beta \cdot \bw_{i}\|_{\ell_1}
\\
\approx&\argmin\limits_{\bw} E(\cdot) + 2\sum_{i=1}^{m} \|\beta \cdot \bw_{i}\|_{\ell_1}
\label{eq:general:summary:rwgLasso}
\end{aligned}
\end{equation}
Combine \eqref{eq:general:summary:rwgLasso} and \eqref{eq:general:summary:cccp-5}, the $\br$ could be calculated.
With Eq~\eqref{eq:general:summary:gammastarupdate2},~\eqref{eq:general:summary:cccp-5} and~\eqref{eq:general:summary:rwgLasso},
the weight ${\bw}$ and hyperparameter ${\bG}$ could be updated alternatively.

In this paper, the $\alpha$ calculated in~\eqref{eq:general:alpha1k} is adopted as the determining factor for connection redundancy.
$\alpha$ is mainly decided by the uncertainty $\Hyper$ and the Hessian. 
Typically, the change of $\alpha$ is the opposite of $\Hyper$.
An increase in $\Hyper$ will cause $\alpha$ to decrease, thereby reducing regularization on corresponding weight $\bw$. 
Based on this, the binary matrix ${C}$ is generated, which denotes the connection redundancy.
$C$ has the same dimension as $\bw$ and will be optimized during the training process. The value is decided by:
\begin{equation}
\label{eq:mask_single_update}
C = \left\{
\begin{array}{ll}
0, &  {\alpha}>\kappa_{\alpha}\\
1, &  \text{others} \\
\end{array}
\right.
\end{equation}
where $\kappa_{\alpha}$ stands for the thresholds for connection pruning. $0$ denotes the redundancy, and $1$ means the weight should be retained. 
It should be noted that the \textit{mask} $C$ will be updated at the last epoch of each cycle.

\subsection{Structured regularization}
\label{subappsec:group_prior_derivation}

For the structured regularization, a group Gaussian prior $p(\bw_g) = \bN(\bw_{g}|\mathbf{0},\bG_{g}) = \prod_{i=1}^{m}\bN(\bw_{gi}|0,\hyper_{g})$ is imposed on a group of weights. It should be noted that the \textit{group} means that connections within a group share the same value on $\hyper$.
The derivation of the loss function for structured regularization is the same as Sec.~\ref{subappsec:non-structured regularization}. 
The only difference is the optimization step for $\hyper_g$:
\begin{equation}
\begin{aligned}
\hyper_g =\frac{\|{\bw_g}\|_{\ell_2}}{\beta_g}= \frac{ \sqrt{\sum\nolimits_{i=1}^m(\bw_i^2)}}{\sqrt{\alpha_g}},  \forall i.
\label{eq:general:summary:cccp-7}
\end{aligned}
\end{equation}
where 
\begin{math}
\alpha_g = \sum\nolimits_{i=1}^{m}(-{\mask_{gi}}/{\hyper_g^2} + 1/{{\hyper_g}}),
 \quad
\mask_{gi} = \left({\hyper_{g}^{-1}}+{\Hessian}({\bw_{gi}}, {\variance}^2)\right)^{-1} 
\end{math}.
For the structured regularization, $\alpha_g$ will be used as the determining factor for group redundancy.
If the value of $\alpha_g$ is smaller than the threshold $\kappa_{\alpha_g}$, the weights of the group will be retained. 
On the contrary, the group's weights will be removed if the value $\alpha_g$ is larger than the threshold $\kappa_{\alpha_g}$.
The binary matrix ${C_g}$ for structured regularization is decided by:
\begin{equation}
\label{eq:mask_group_update}
C_g = \left\{
\begin{array}{ll}
0, &  {\alpha_g}>\kappa_{\alpha_g}\\
1, &  \text{others} \\
\end{array}
\right.
\end{equation}
\textit{mask} $C_g$  has the same same dimension as $\bw$ and will be updated at the last epoch of each cycle as well. 
\section{Experiment}
\label{sec:app_result}
The algorithm is demonstrated on the chaotic Lorenz system~\cite{lorenz1963deterministic}, Lotka-Volterra system~\cite{berryman1992orgins} and Kolmogorov–Petrovsky–Piskunov (Fisher-KPP) system~\cite{fisher1937wave}.
All experiments are performed in PyTorch framework by using a single GPU (NVIDIA TITAN V). The code is available on~\url{https://github.com/nips2021anonymous/MathONet}.
 
\subsection{Chaotic Lorenz System}
\label{subsec:app_result_lorenz}

\subsubsection{Experiment Setup}
\label{appsubsubsec:Lorenz_detials}
The MathONet is initialized with $1$ hidden layer and $3$ hidden neurons. 
The OperNet includes $5$ basic unary functions, i.e.~\texttt{identity}, \texttt{sin}, \texttt{cos}, \texttt{log}, \texttt{exp}. 
The initial value of the regularization parameters $\lambda, \lambda_g$ are assigned from the set $\{1e^{-2}, 1e^{-4}, 1e^{-6}, 1e^{-8}, 1e^{-10}\}$ and are decayed to one-tenth every $200$ epochs. For each hyper-parameter, the identification procedure is repeated
every $200$ epochs. For each hyper-parameter, the identification procedure is repeated $10$ times with differing weight initialization. 
\subsubsection{Result on Noise-free Dataset}
\begin{figure}
	\centering
	\begin{subfigure}[b]{0.44\textwidth}
	\centering
	\includegraphics[height = 2.6cm]{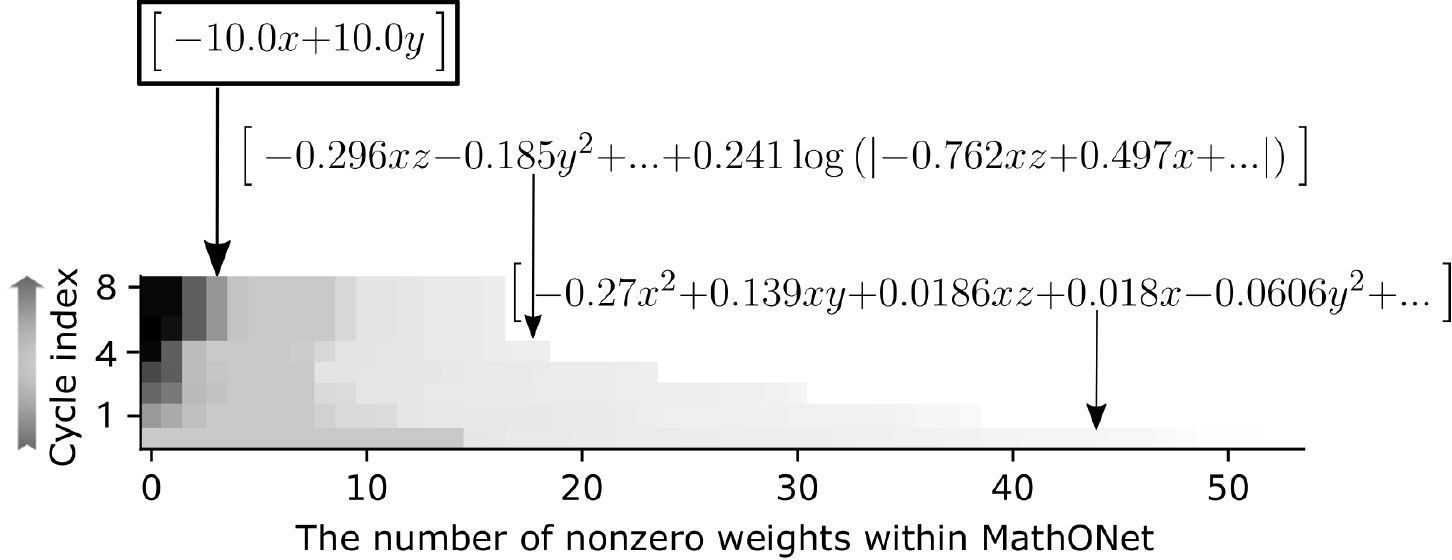}
	\caption{The number of nonzero weights of the MathONet in each cycle.}
	\label{fig:lorenz_weight_equation_change_model_x}
\end{subfigure}
\hspace{0.1\textwidth}
\begin{subfigure}[b]{0.44\textwidth}
	\centering
	\includegraphics[height = 3.2cm]{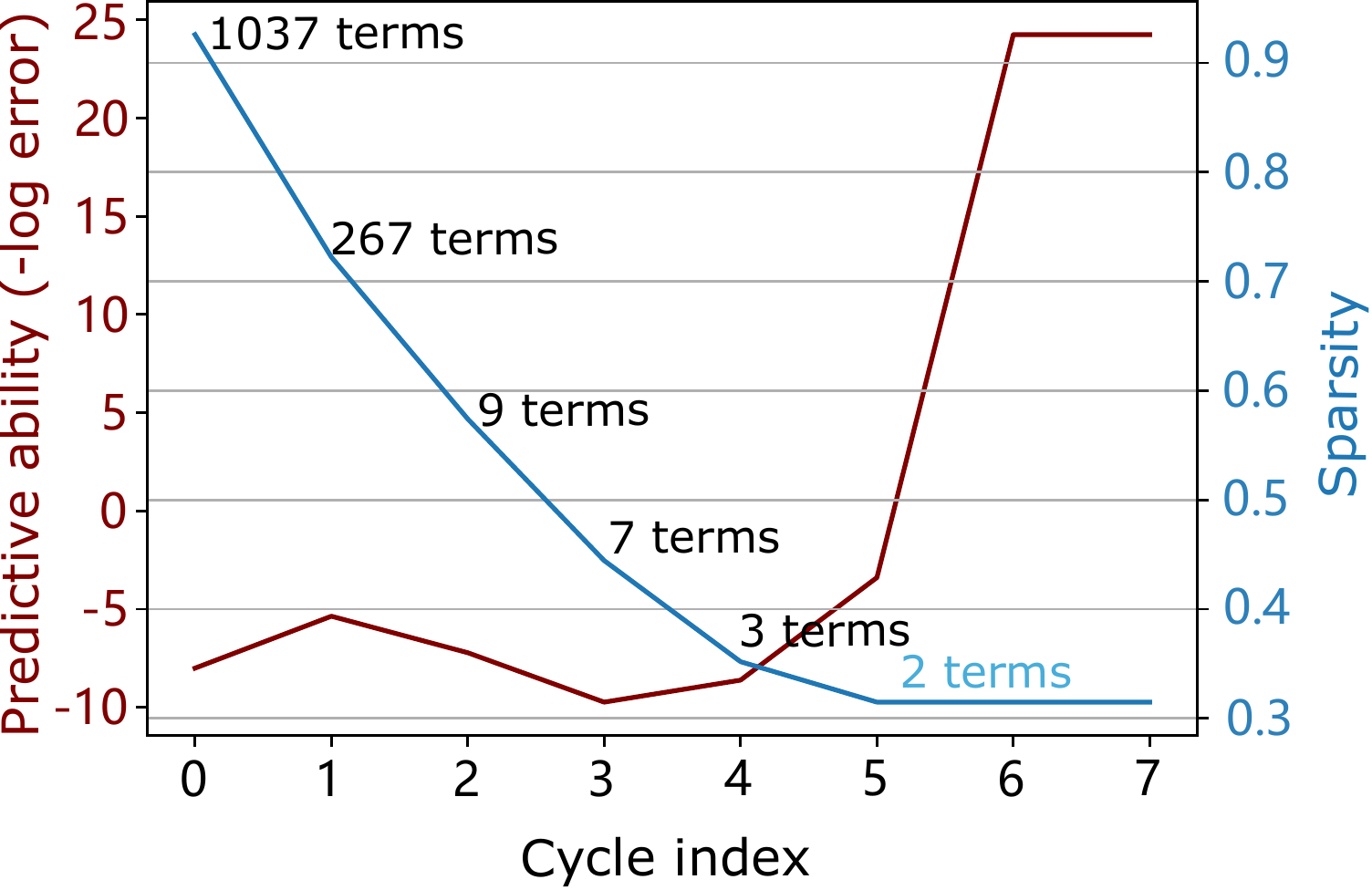}
	\caption{The sparsity and predictive ability of the MathONet in each cycle.}
	\label{fig:lorenz_sparsity_loss_change_model_x}
\end{subfigure}   
	\caption{ 
The sparsity, predictive ability and weights of the identified MathONet generated in each cycle, which aims to discover governing equation of model $x$ of Lorenz system.
a) The nonzero weights of the MathONet generated in each cycle. 
The horizontal axis represents the combination of non-zero weights in the MathONet generated in each cycle.
The vertical axis denotes the index of training cycles.
The expression at each turning line (the cliff) represents the governing equation identified in the corresponding cycle.
b) The sparsity and prediction ability of the MathONet identified in each cycle. The model becomes more and more sparse and has more and more predictive ability. The annotation next to the sparsity line represents the number of identified mathematical terms of the corresponding cycle. The first cycle represents the result identified by sparse group Lasso method which is still redundant (267 terms), and the prediction ability is low. The correct structure and coefficients can be identified around $7$ cycles.   	
}
	\label{fig:lorenz_weight_sparsity_loss_change_model_x}
\end{figure}

\begin{figure}
	\centering
	\begin{subfigure}[b]{0.44\textwidth}
	\centering
	\includegraphics[height = 2.8cm]{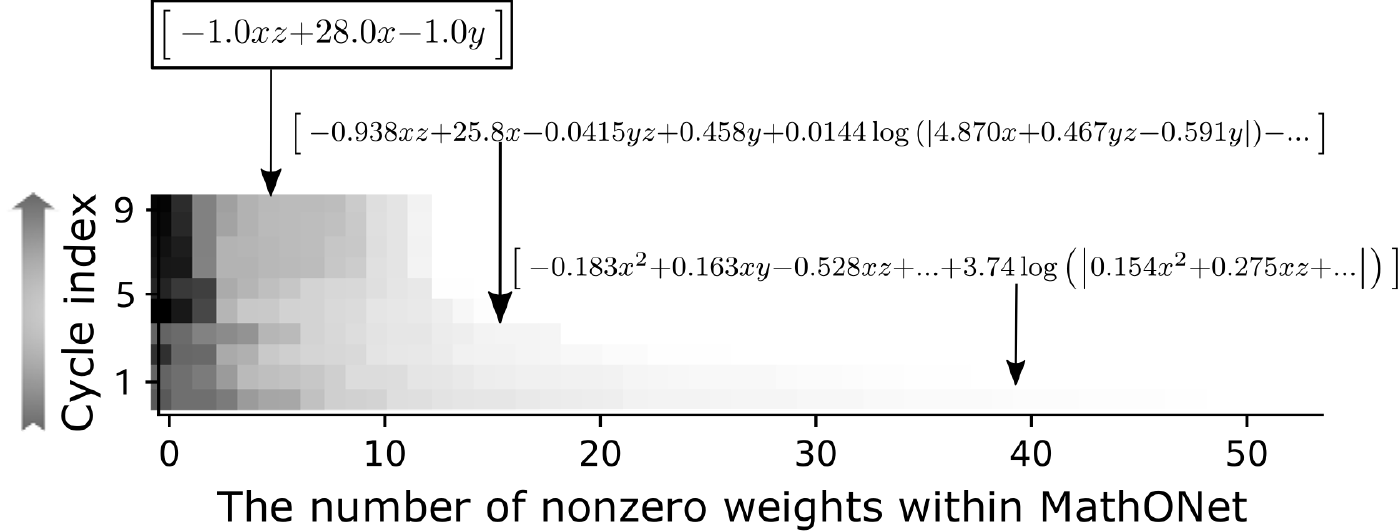}
	\caption{The number of nonzero weights of the MathONet in each cycle.}
	\label{fig:lorenz_weight_equation_change_model_y}
\end{subfigure}
\hspace{0.1\textwidth}
\begin{subfigure}[b]{0.44\textwidth}
	\centering
	\includegraphics[height = 3.4cm]{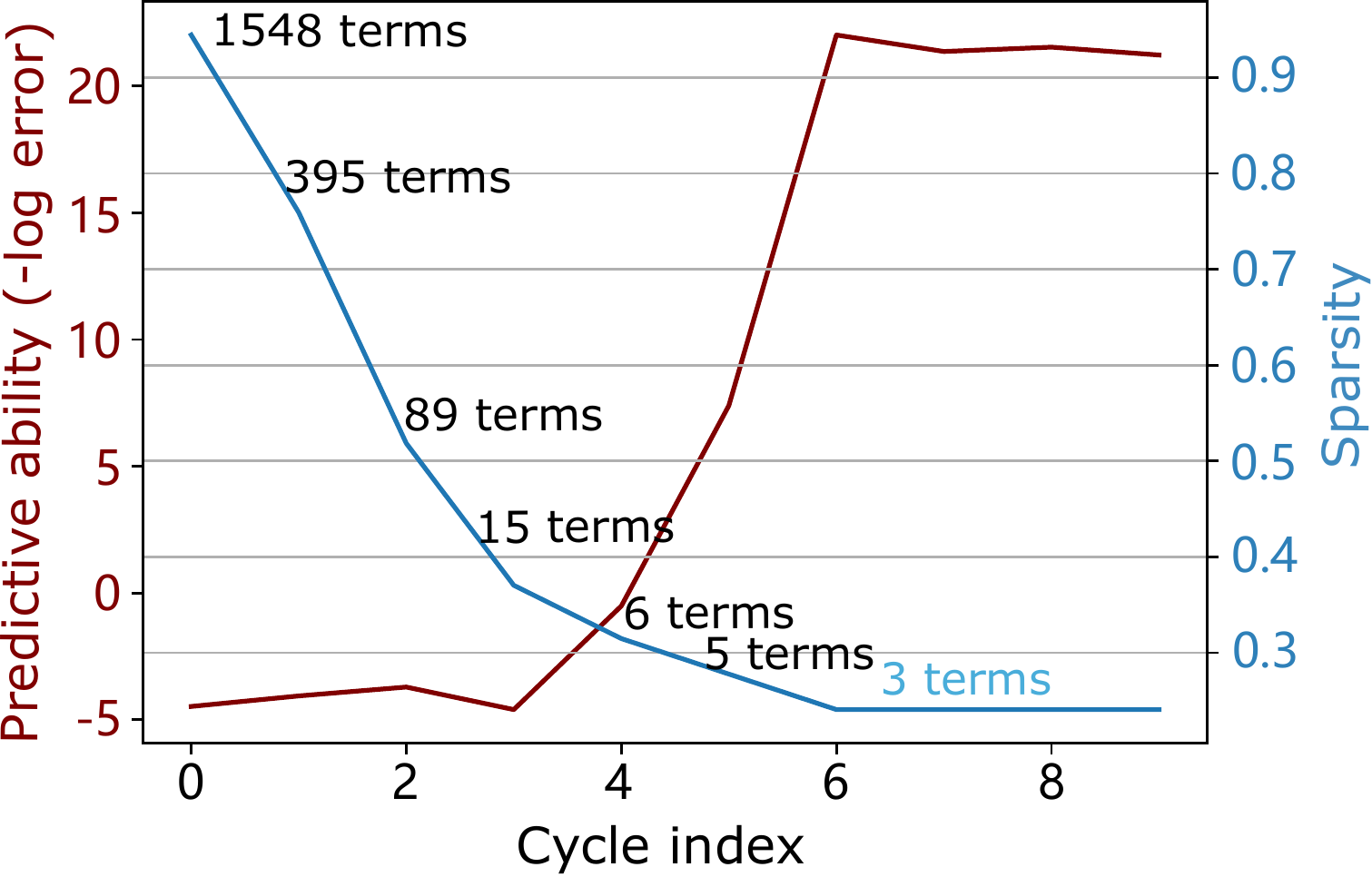}
	\caption{The sparsity and predictive ability of the MathONet in each cycle.}
	\label{fig:lorenz_sparsity_loss_change_model_y}
\end{subfigure}   
	\caption{The sparsity, predictive ability and weights of the identified MathONet generated in each cycle, which aims to discover governing equation of model $y$ of Lorenz system.
		a) The nonzero weights of the MathONet generated in each cycle. 
		The horizontal axis represents the combination of non-zero weights in the MathONet generated in each cycle.
		The vertical axis denotes the index of training cycles.
		The expression at each turning line (the cliff) represents the governing equation identified in the corresponding cycle.
		b) The sparsity and prediction ability of the MathONet identified in each cycle. The model becomes more and more sparse and has more and more predictive ability. The annotation next to the sparsity line represents the number of identified mathematical terms of the corresponding cycle. The first cycle represents the result identified by sparse group Lasso method which is still redundant ($395$ terms), and the prediction ability is low. The correct structure and coefficients can be identified around $9$ cycles. 
	}
	\label{fig:lorenz_weight_sparsity_loss_change_model_y}
\end{figure}

With the simulated dataset without noise, the identified governing equations without fine-tuning are:
\begin{subequations}
	\begin{align}
	&\Dot{x} = -10.000x + 10.000y
	\\
	&\Dot{y} = -1.000xz + 28.000x - 1.000y
	\\
	&\Dot{z} = 1.000xy - 2.667z
	\end{align}
	\label{eq:identified lorenz}
\end{subequations}
It can be observed that both the equation form and parameters are captured accurately. 
Fig.~\ref{fig:lorenz_weight_sparsity_loss_change_model_x} and Fig.~\ref{fig:lorenz_weight_sparsity_loss_change_model_y} show the process of the algorithm searching for the governing equations for model ${x}$ and  ${y}$, respectively. 
In Fig.~\ref{fig:lorenz_weight_equation_change_model_x} and Fig.~\ref{fig:lorenz_weight_equation_change_model_y}, the non-zero weight elements within the MathONet of each cycle are collected to form a weight vector. 
Fig.~\ref{fig:lorenz_sparsity_loss_change_model_x} and Fig.~\ref{fig:lorenz_sparsity_loss_change_model_y} show that the predictive ability is improved as the model complexity decreases.
It only takes $7$ cycles to identify the correct structures and coefficients for model ${x}$ and $9$ cycles for model ${y}$. This is reasonable because the model ${y}$ is more complex with a second-order term than model ${x}$ as described in~\eqref{eq:identified lorenz}.

The predicted distribution of the identified model is also investigated.
We sampled a total of 1000 times based on the identified model and parameter. 
The first row in Fig.~\ref{fig:lorenz_uncertainty} shows the predicted mean and variance of each model on the dataset without noise. 
Although the variance is very small almost for all data points, 
it still can be observed that the predicted distribution spreads a bigger range around the turning points, which means more training data is required around these points. 

\subsubsection{Result on Noisy Dataset}
\begin{figure}
	\centering
	\begin{subfigure}[b]{0.3\textwidth}
		\centering
		\includegraphics[width=\textwidth]{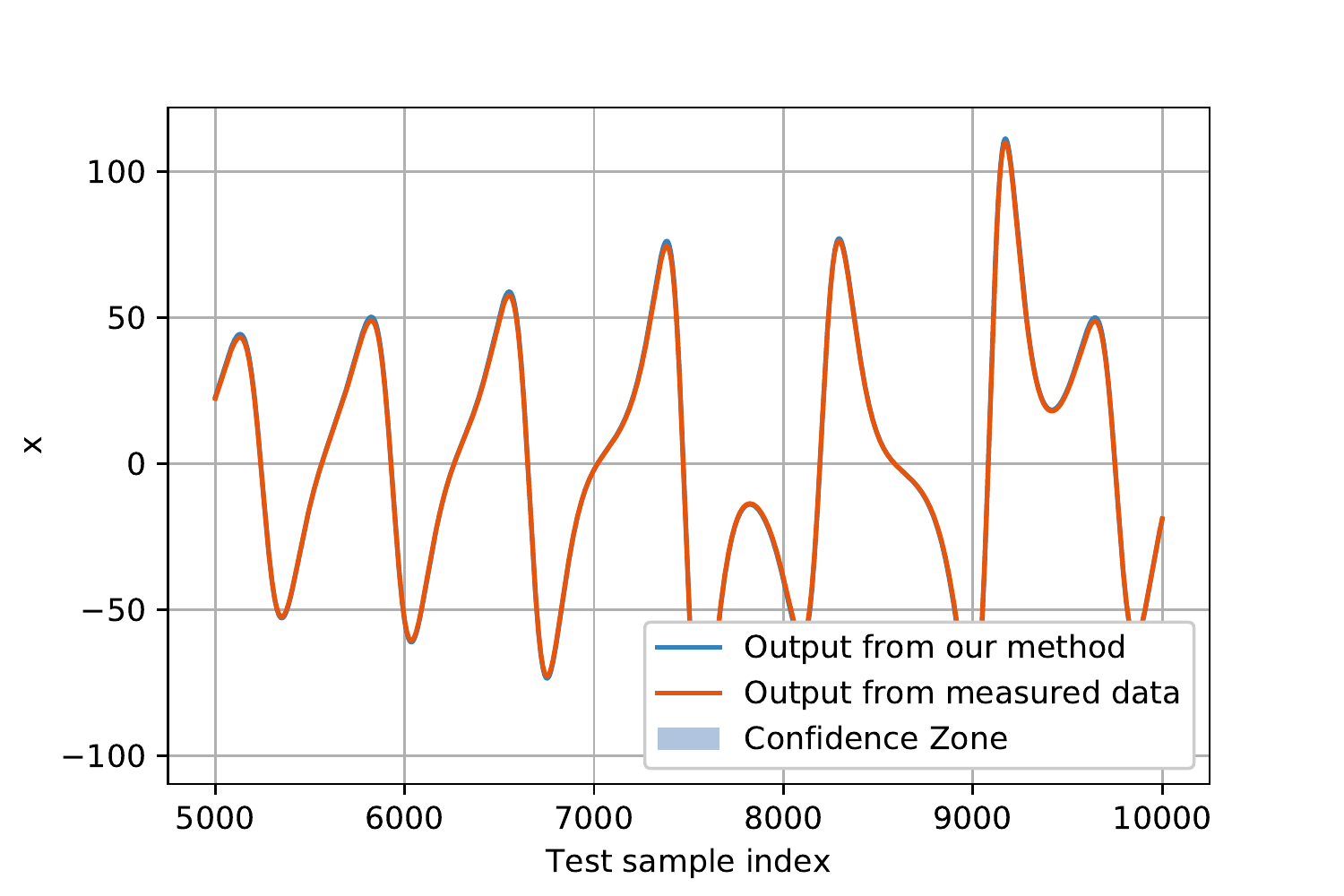}
		\caption{Model $x$, $\sigma = 0$}
		\label{fig:lorenz_uncertainty_model_x_noise_0_zoom}
	\end{subfigure}
	\hfill
	\begin{subfigure}[b]{0.3\textwidth}
		\centering
		\includegraphics[width=\textwidth]{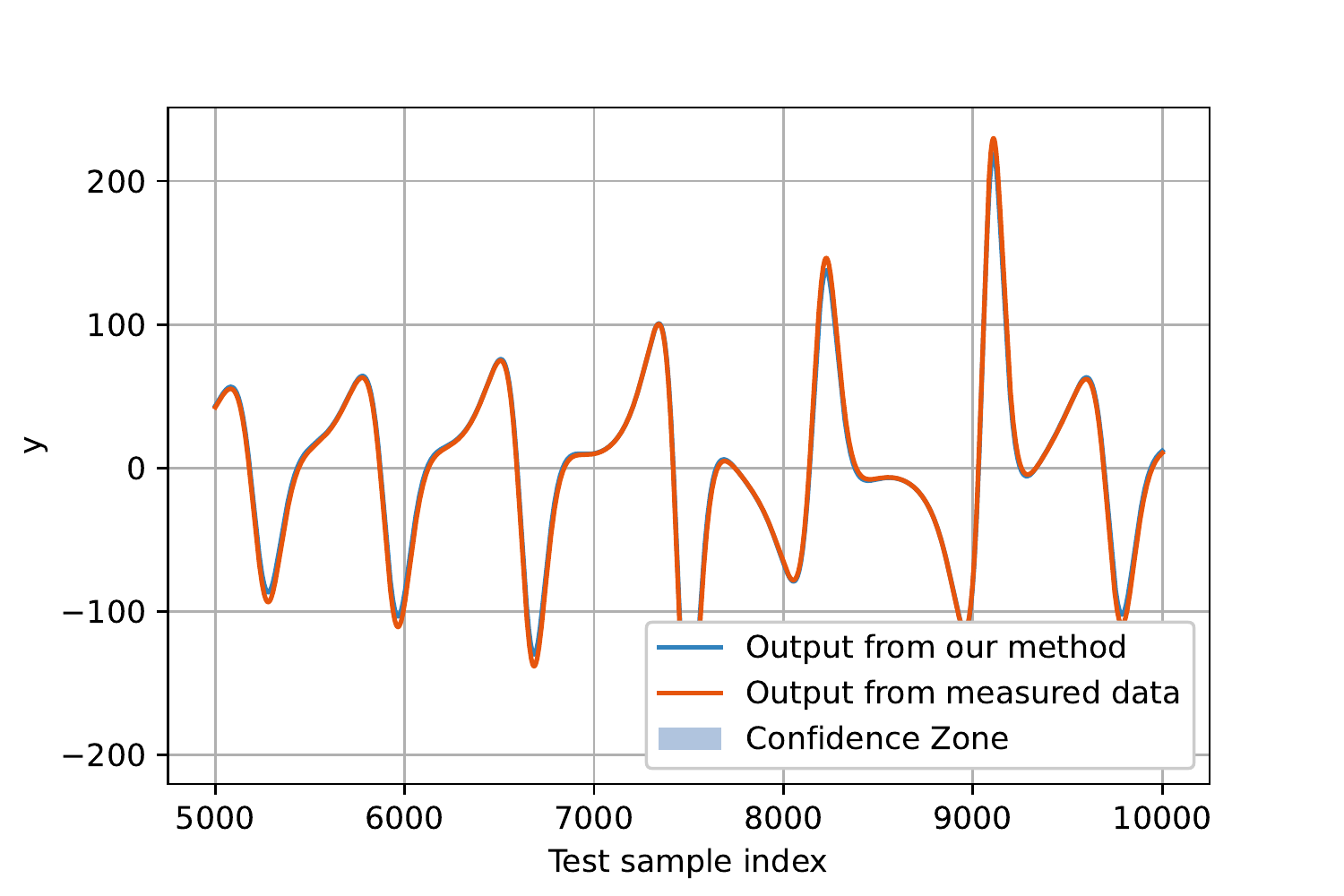}
		\caption{Model $y$, $\sigma = 0$}
		\label{fig:lorenz_uncertainty_model_y_noise_0_zoom}
	\end{subfigure}
	\hfill 
	\begin{subfigure}[b]{0.3\textwidth}
		\centering
		\includegraphics[width=\textwidth]{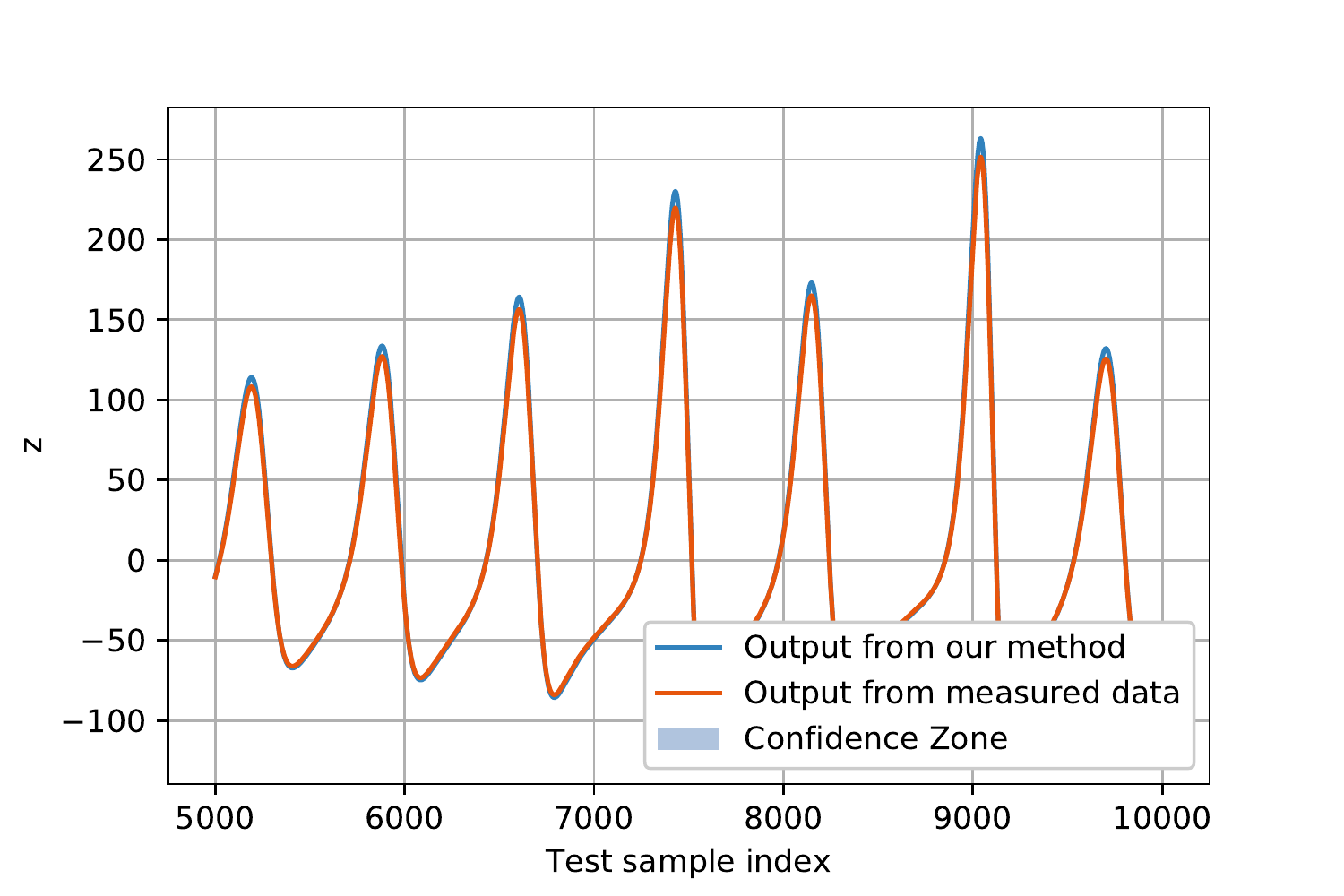}
		\caption{Model $z$, $\sigma = 0$}
		\label{fig:lorenz_uncertainty_model_z_noise_0_zoom}
	\end{subfigure}
	\\
	\begin{subfigure}[b]{0.3\textwidth}
		\centering
		\includegraphics[width=\textwidth]{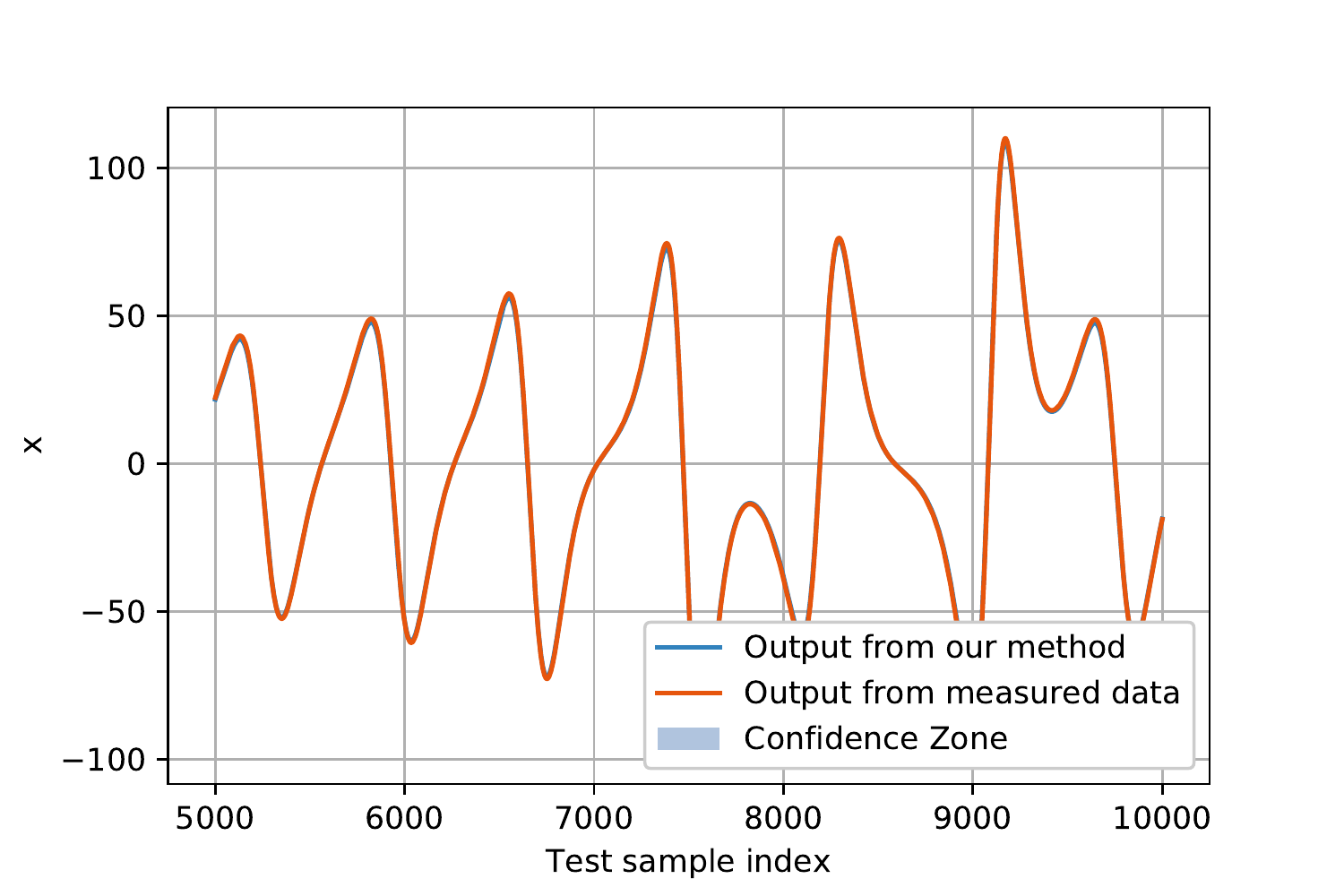}
		\caption{Model $x$, $\sigma = 0.01$}
		\label{fig:lorenz_uncertainty_model_x_noise_001_zoom}
	\end{subfigure}
	\hfill
	\begin{subfigure}[b]{0.3\textwidth}
		\centering
		\includegraphics[width=\textwidth]{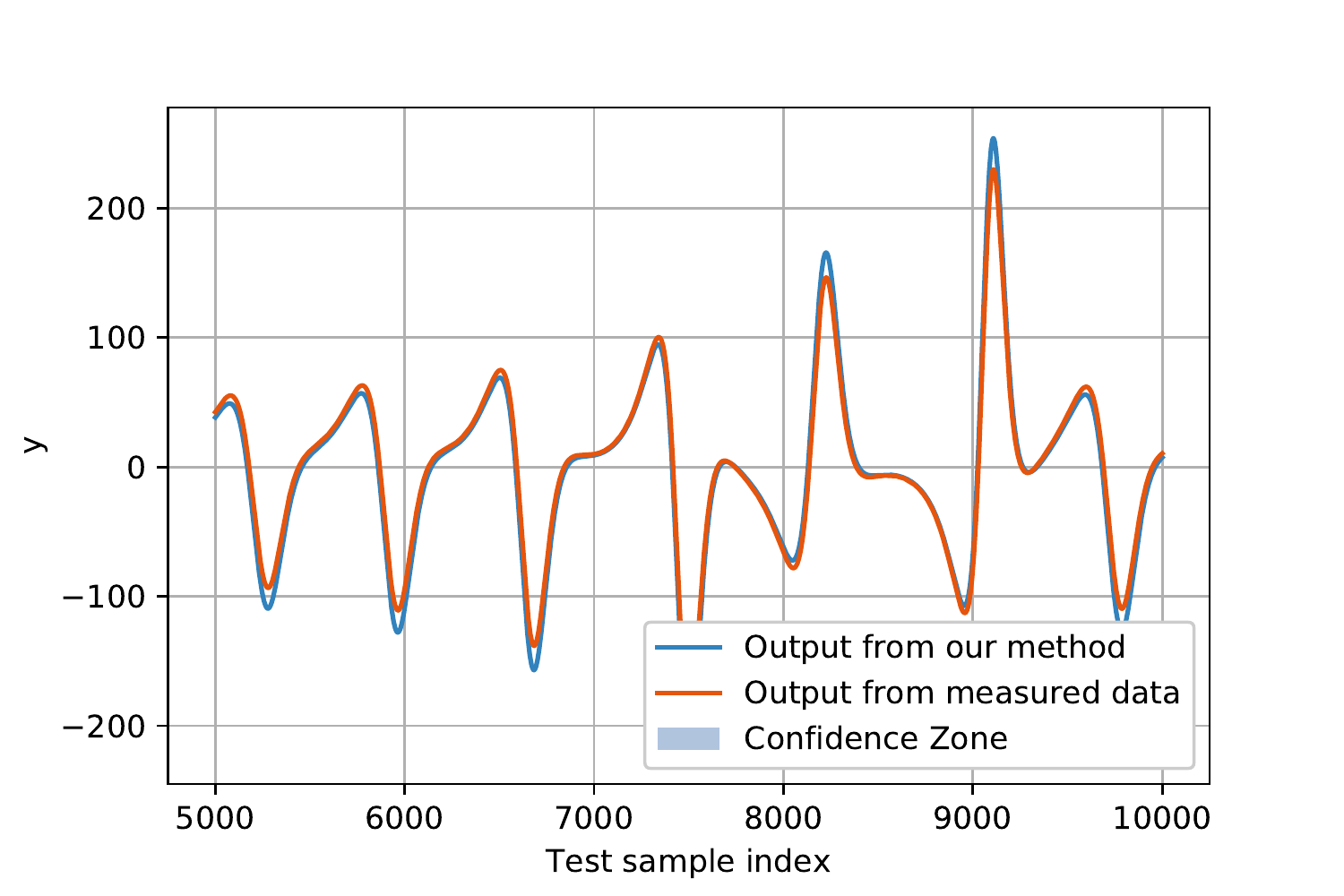}
		\caption{Model $y$, $\sigma = 0.01$}
		\label{fig:lorenz_uncertainty_model_y_noise_001_zoom}
	\end{subfigure}
	\hfill 
	\begin{subfigure}[b]{0.3\textwidth}
		\centering
		\includegraphics[width=\textwidth]{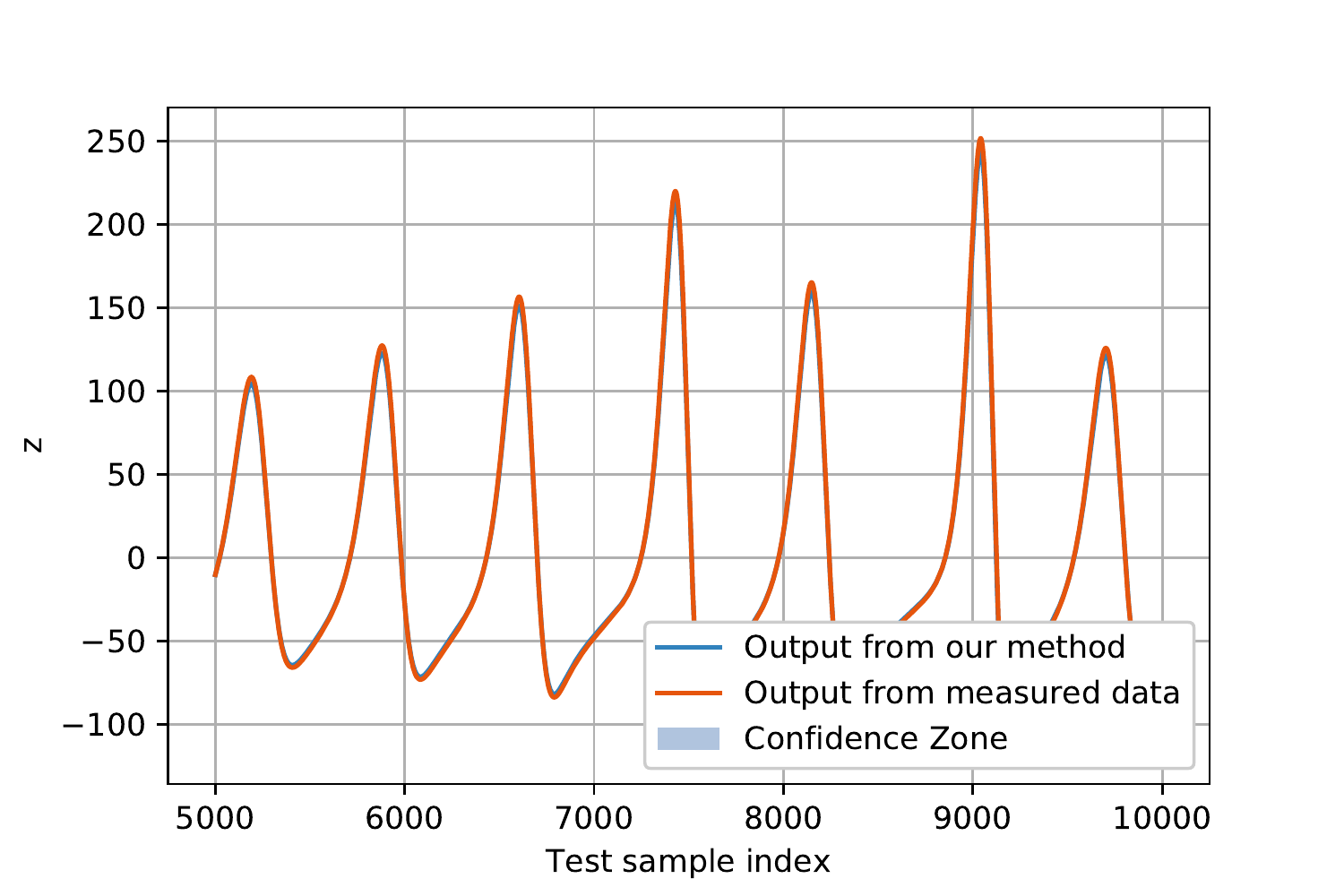}
		\caption{Model $z$, $\sigma = 0.01$}
		\label{fig:lorenz_uncertainty_model_z_noise_001_zoom}
	\end{subfigure}   
	\\
	\begin{subfigure}[b]{0.3\textwidth}
		\centering
		\includegraphics[width=\textwidth]{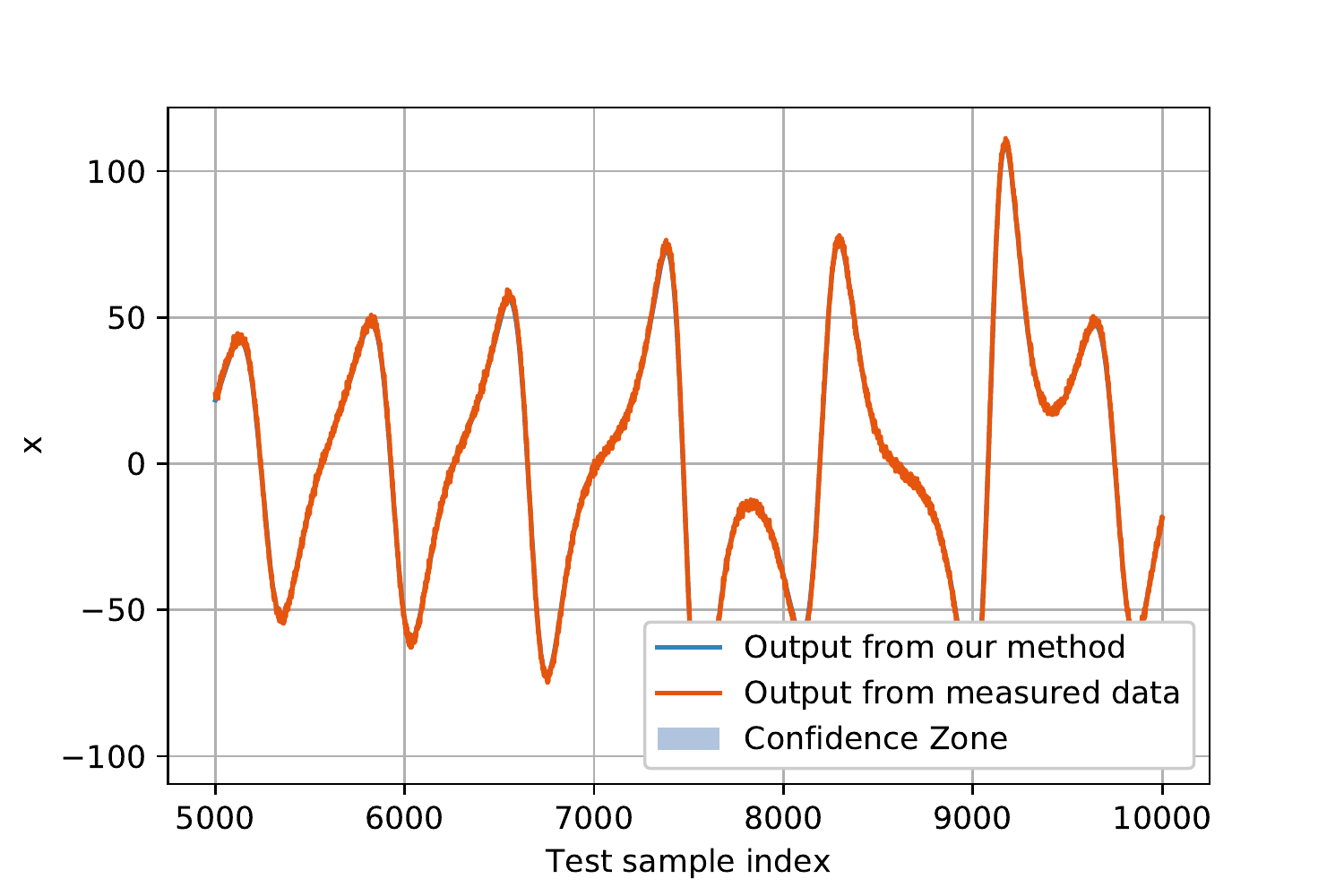}
		\caption{Model $x$, $\sigma = 1$}
		\label{fig:lorenz_uncertainty_model_x_noise_1_zoom}
	\end{subfigure}
	\hfill
	\begin{subfigure}[b]{0.3\textwidth}
		\centering
		\includegraphics[width=\textwidth]{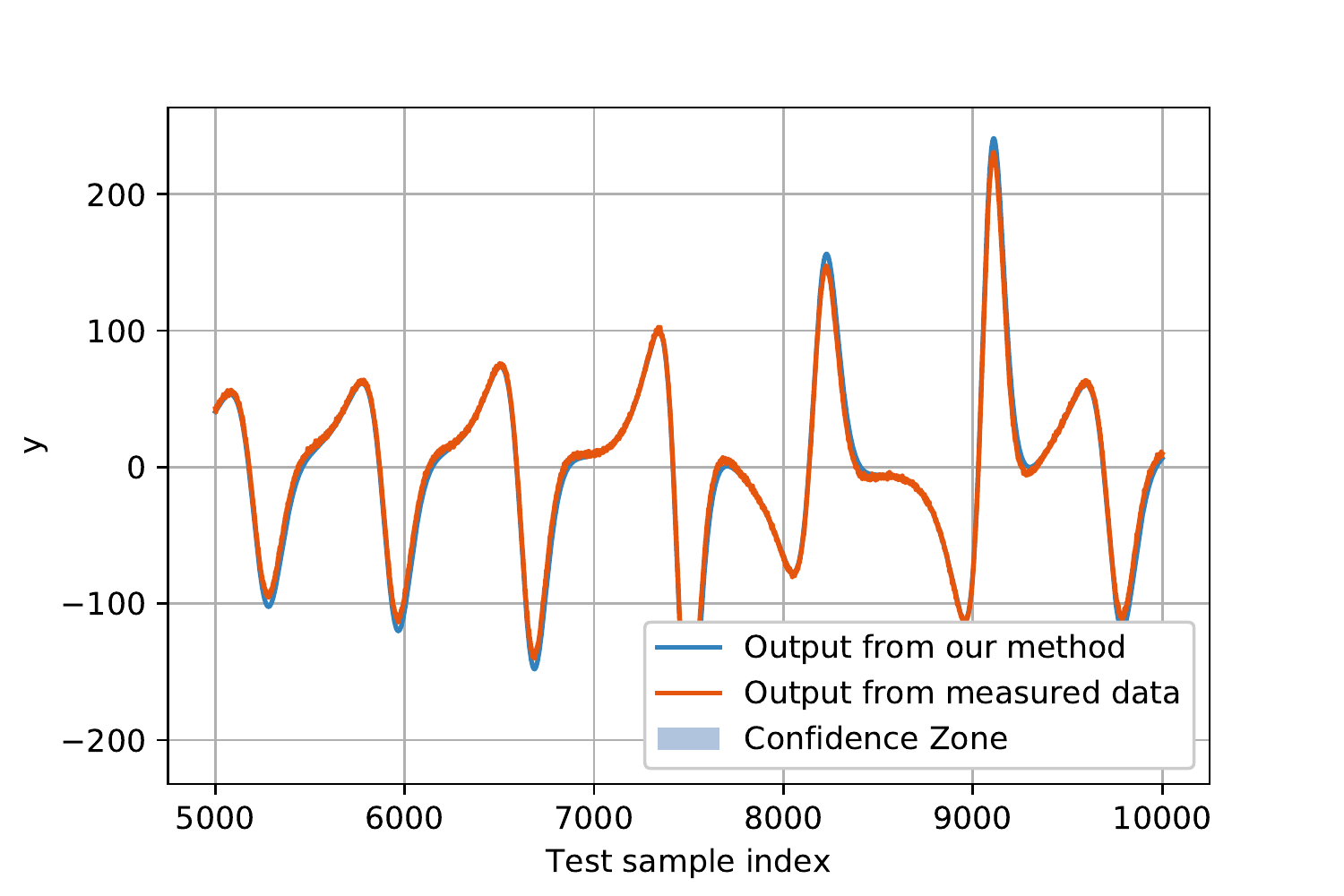}
		\caption{Model $y$, $\sigma = 1$}
		\label{fig:lorenz_uncertainty_model_y_noise_1_zoom}
	\end{subfigure}
	\hfill 
	\begin{subfigure}[b]{0.3\textwidth}
		\centering
		\includegraphics[width=\textwidth]{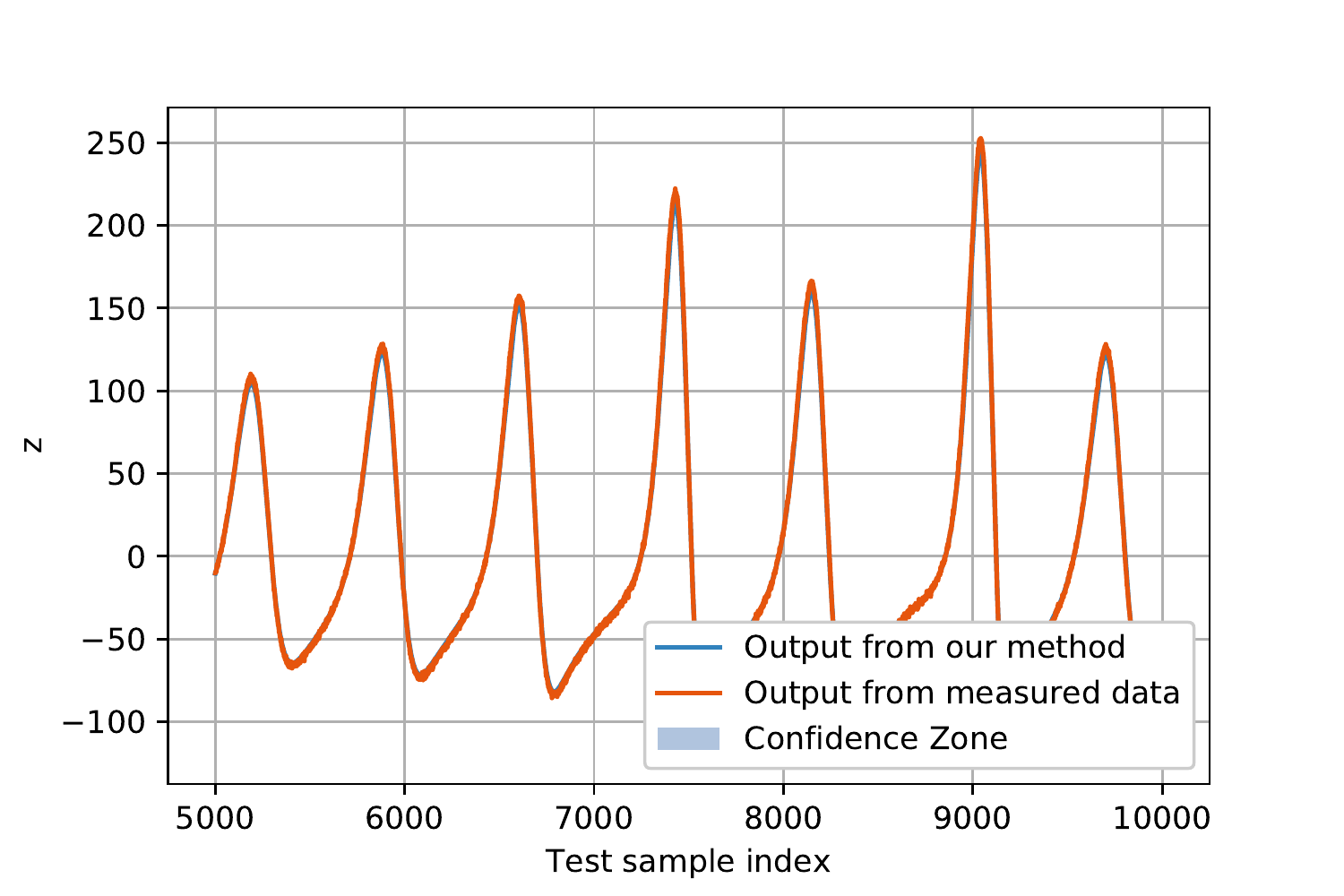}
		\caption{Model $z$, $\sigma = 1$}
		\label{fig:lorenz_uncertainty_model_z_noise_1_zoom}
	\end{subfigure}        
	\\
	\begin{subfigure}[b]{0.3\textwidth}
		\centering
		\includegraphics[width=\textwidth]{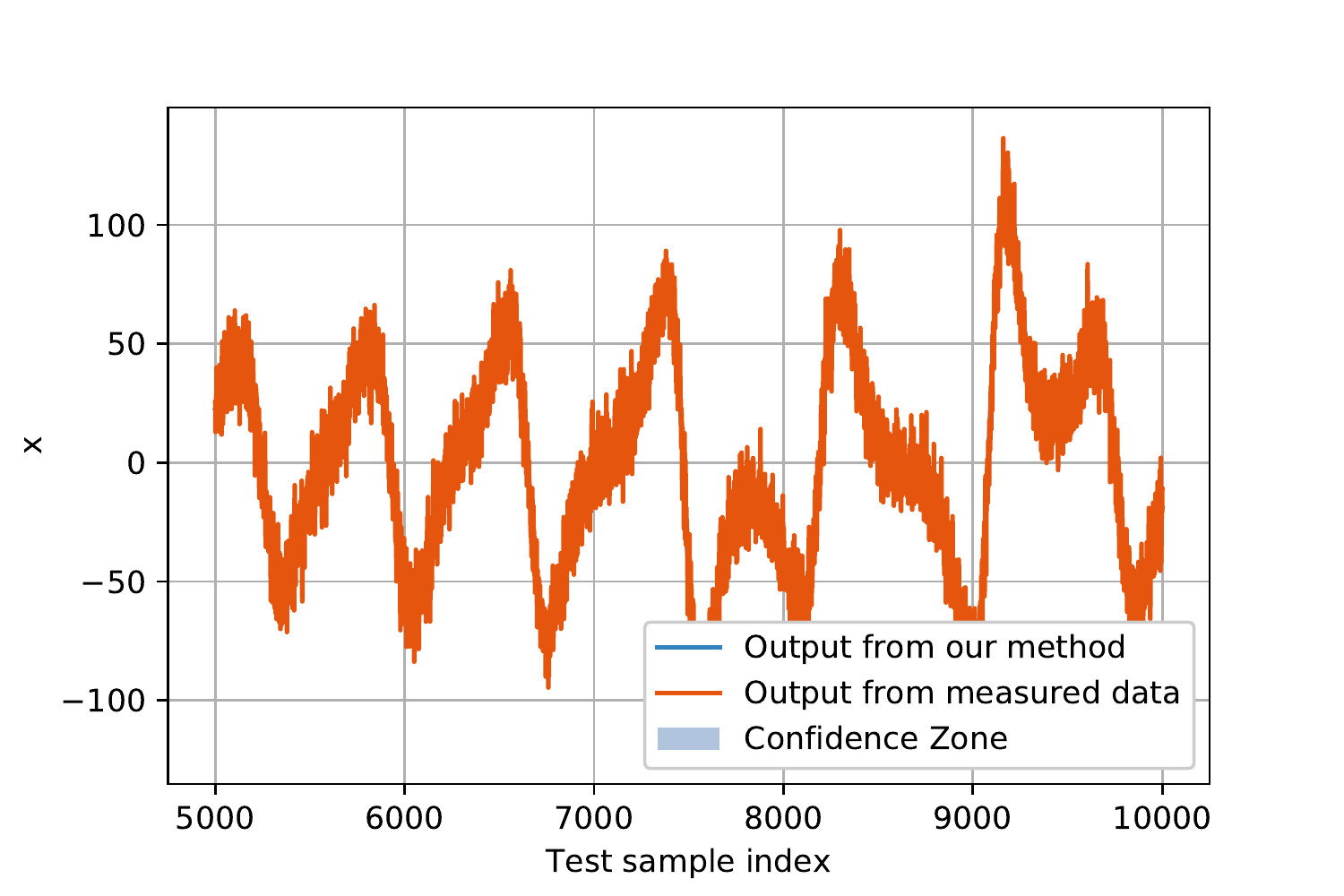}
		\caption{Model $x$, $\sigma = 10$}
		\label{fig:lorenz_uncertainty_model_x_noise_10_zoom}
	\end{subfigure}
	\hfill
	\begin{subfigure}[b]{0.3\textwidth}
		\centering
		\includegraphics[width=\textwidth]{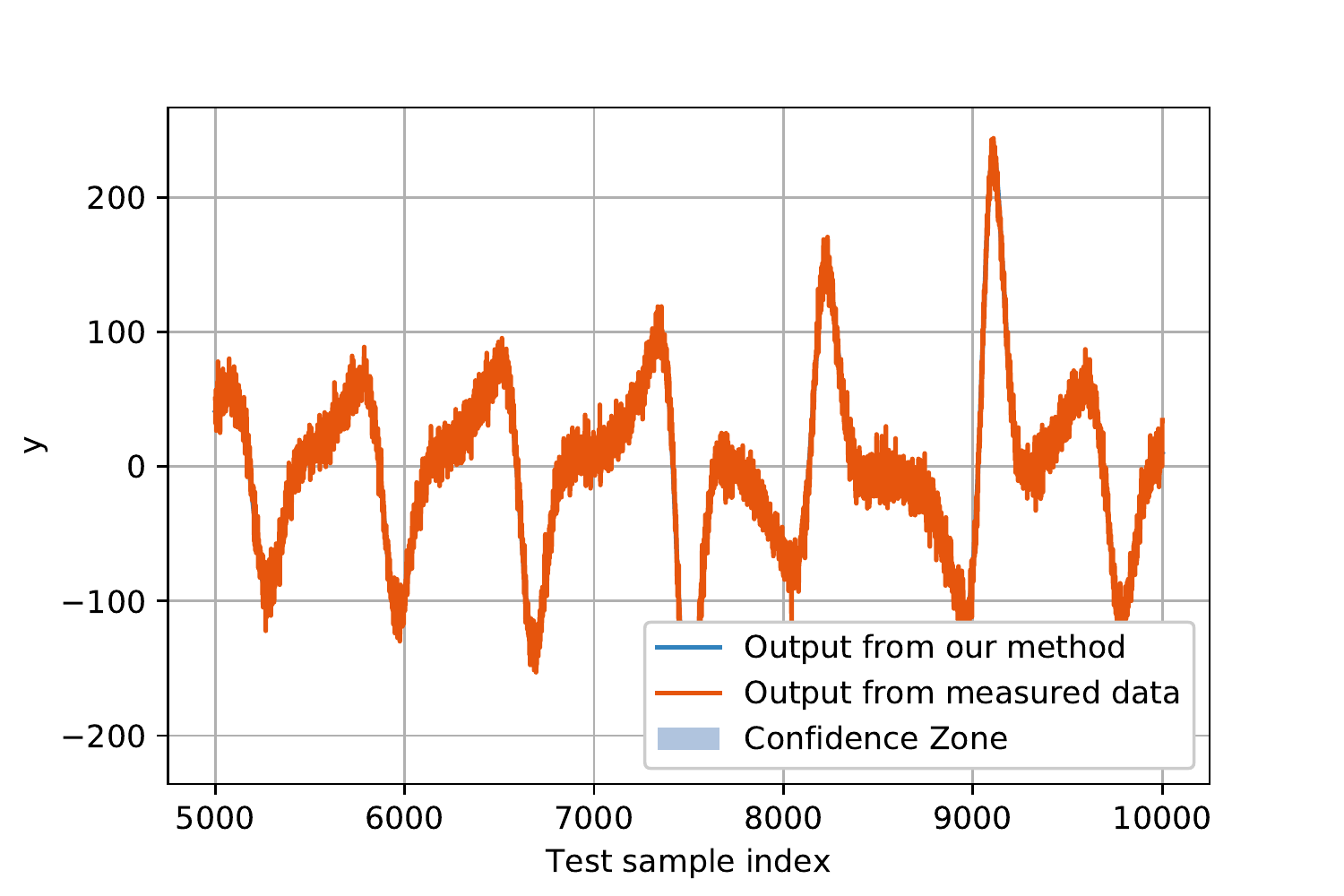}
		\caption{Model $y$, $\sigma = 10$}
		\label{fig:lorenz_uncertainty_model_y_noise_10_zoom}
	\end{subfigure}
	\hfill 
	\begin{subfigure}[b]{0.3\textwidth}
		\centering
		\includegraphics[width=\textwidth]{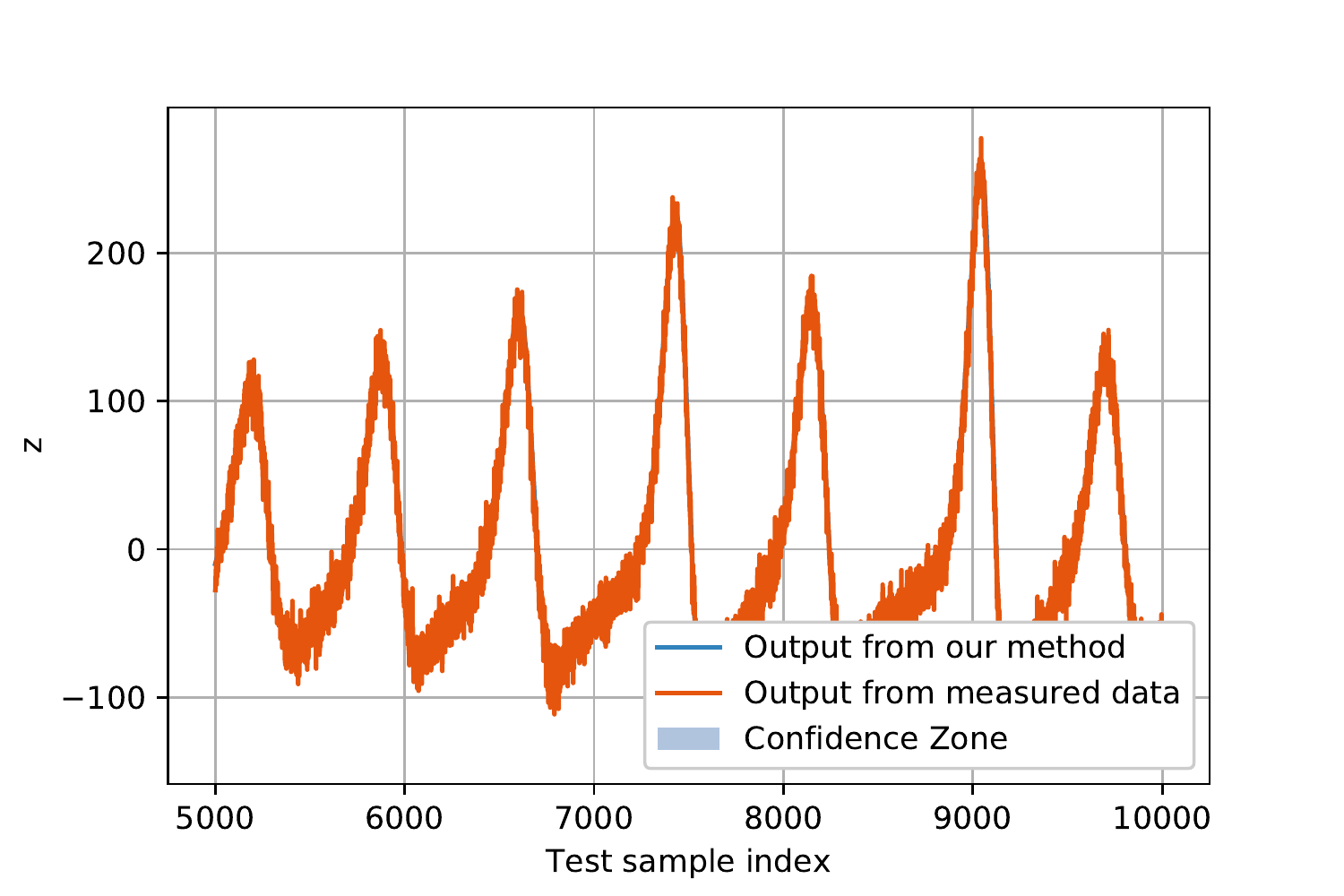}
		\caption{Model $z$, $\sigma = 10$}
		\label{fig:lorenz_uncertainty_model_z_noise_10_zoom}
	\end{subfigure}        
	\caption{Predicted distribution for the identified Lorenz system generated from data with different noise $\noise = \prob(0,\variance^2)$, where $\variance \in \{0.01, 1, 10\}$.The blue curve stands for the 
		best-predicted output of our method. The shaded area 
		represents the model uncertainty (showing $2$ standard deviations)}
	\label{fig:lorenz_uncertainty}
\end{figure}

We also test the robustness of the algorithm with noisy derivatives measurement, Gaussian noise $\noise = \prob(0,\variance^2)$ with $\variance \in \{0.01, 1, 10\}$ is added to the exact derivatives, respectively. 
The experiments were implemented with the same experiments setting as the simulated dataset in the Sec.~\ref{sebsec:Lorenz System}.
As shown in Table~\ref{tab:lorenz_noisy_result}, both structure and parameters are correctly identified even under the large noise value ($\variance = 10$). 
The coefficients $\sigma, \beta, \gamma$ can be determined accurately within $0.5\%$ around the true value. 
Fig.~\ref{fig:lorenz_uncertainty} is the prediction uncertainty for each dataset, which shows the predicted uncertainty of the identified model improves along with increasing $\variance$.

\begin{table}[]
	\caption{The identified governing equations for the chaotic Lorenz system with noisy measurements
	}
	\centering
	\label{tab:lorenz_noisy_result}
	\begin{tabular}{l|l}
		\hline
		Noise ($\variance)$ & \multicolumn{1}{c}{Identified governing equations}  \\ \hline
		\multicolumn{1}{c|}{$0$}    & \begin{tabular}[c]{@{}l@{}}$\Dot{x} = -9.99999935280567x + 9.99999935280567y$ \\ $\Dot{y} = -0.999997263120259xz + 27.9998710115329x - 0.999953545630988y$\\ $\Dot{z} = 0.99999998781167xy - 2.66666669723636z$\end{tabular}                   \\ \hline
		\multicolumn{1}{c|}{$0.1$}  & \begin{tabular}[c]{@{}l@{}}$\Dot{x} = -9.99999496925967x + 9.99999952895684y$\\ $\Dot{y} = -1.00052719172829xz + 28.0258380824985x - 1.00998669015022y$\\ $\Dot{z} = 0.999999113716832xy - 2.66666544271354z$\end{tabular}                    \\ \hline
		\multicolumn{1}{c|}{$1$}    & \begin{tabular}[c]{@{}l@{}}$\Dot{x} = -10.0033841069518x + 10.0034298967151y$\\ $\Dot{y} = -1.00010106902181xz + 28.0040216562134x - 1.00069798536041y$
			\\ $\Dot{z} = 0.999929109879073xy - 2.66622599261675z$\end{tabular}                    \\ \hline
		\multicolumn{1}{c|}{$10$}   & \begin{tabular}[c]{@{}l@{}}$\Dot{x} = -9.99893134664084x + 10.0048109236265y$\\ $\Dot{y} = -1.0092184658872xz + 28.4682092943604x - 1.1929184472878y$\\ $\Dot{z} = 1.00051935274034xy - 2.66709953162919z$\end{tabular}                       \\ \hline
	\end{tabular}
\end{table}

\subsection{Lotka-Volterra System}
\label{appsubsec:app_result_Lotka-Volterra}
\subsubsection{Experiment Setup}
\label{appsubsubsec:Lotka-Volterra_detials}
\begin{figure}
	\centering
	\begin{subfigure}[b]{0.31\textwidth}
		\centering
		\includegraphics[width=\textwidth]{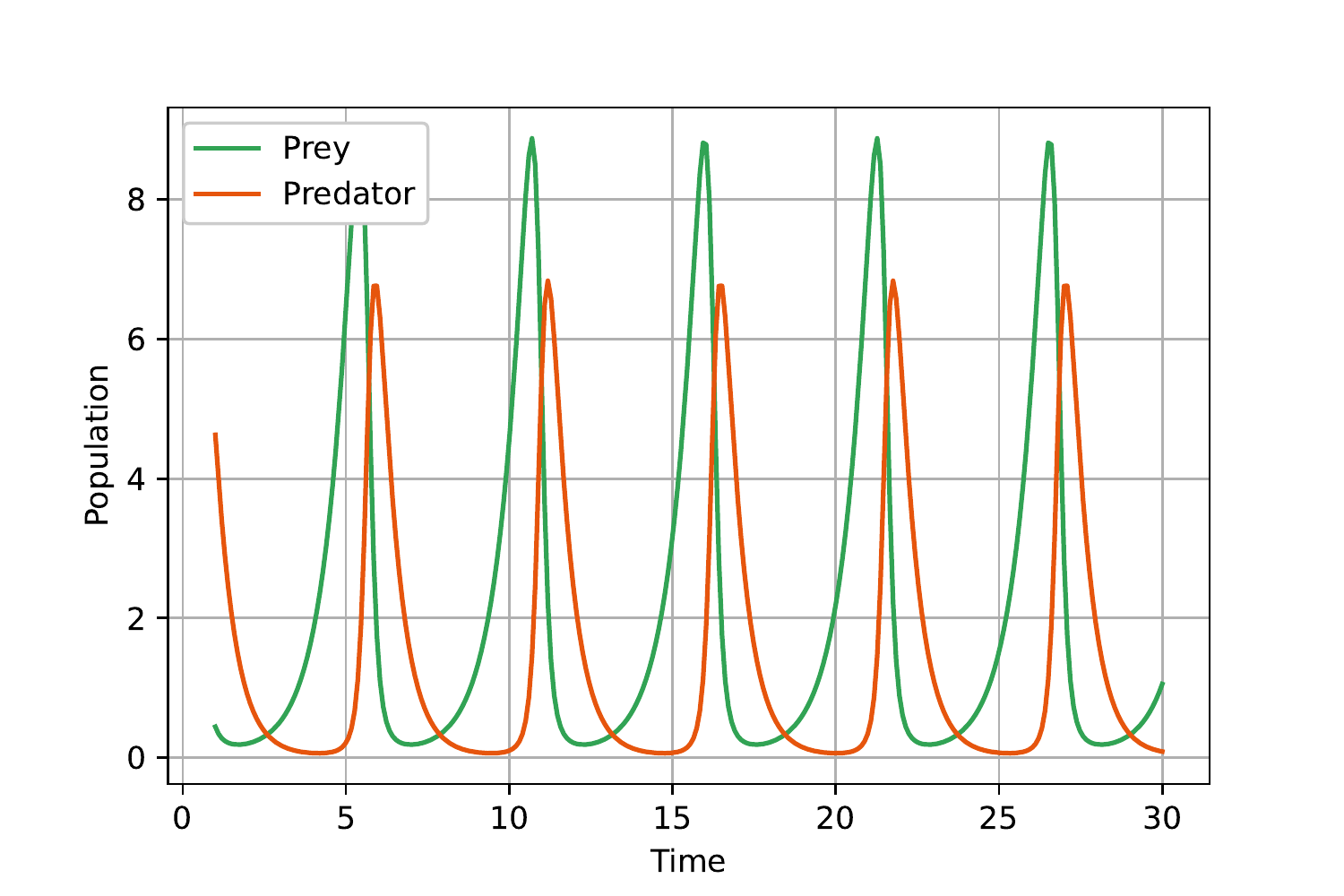}
		\caption{Predator and prey population}
		\label{fig:lk_data_pattern}
	\end{subfigure}
	\begin{subfigure}[b]{0.31\textwidth}
		\centering
		\includegraphics[width=\textwidth]{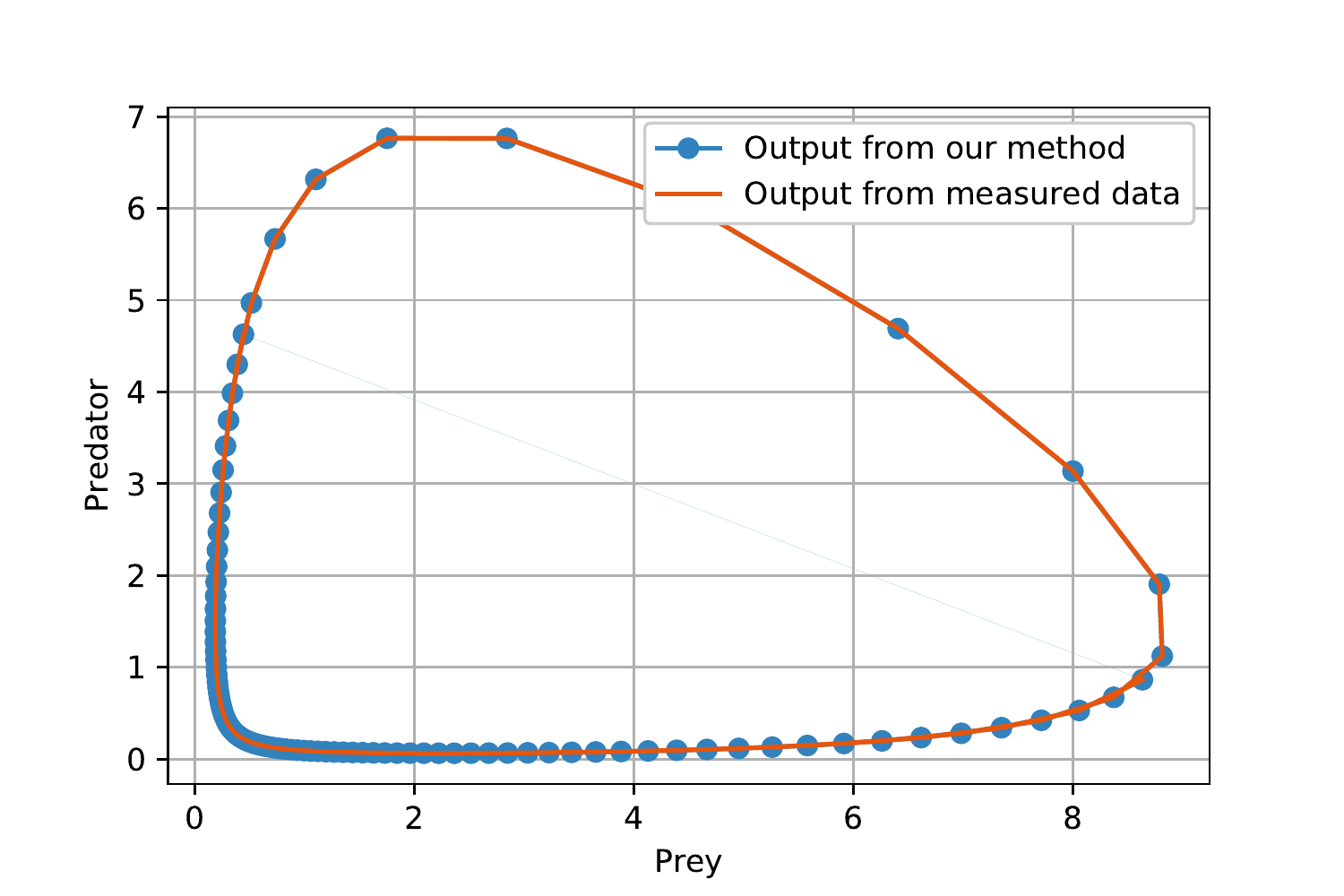}
		\caption{Phase-space plot of solution.}
		\label{fig:lk_solution_contour}
	\end{subfigure}
	\begin{subfigure}[b]{0.31\textwidth}
		\centering
		\includegraphics[scale = 0.1, width=\textwidth]{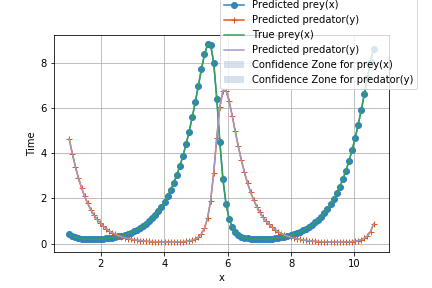}
		\caption{Predicted output. 
		}
		\label{fig:lk_solution_pattern}
	\end{subfigure}
	\caption{The solution of the prey and predator population with the initial conditions $x(0) = 0.442, y(0) = 4.628$. a) The data of the prey and predator population generated by~\eqref{eq:theoretical Lotka Volterra}; b) Phase-space plot for the predator and prey population of the identified model~\eqref{eq:identified Lotka Volterra}; c) The uncertainty of the predicted output. 
	}
	\label{fig:lk_solution}
\end{figure}

The coefficients $\alpha, \beta, \delta, \gamma$ in~Eq.\ref{eq:theoretical Lotka Volterra} are the positive real parameters identified in this experiment and set as $1.3, 0.9, 0.8, 1.8$, respectively. 
Fig.~\ref{fig:lk_data_pattern} shows the generated prey and predator population along time. Specifically, Fig.~\ref{fig:lk_solution_contour} illustrates the dynamic changes of the prey and predator population in a circle of growth and decline.
The input and output data with the dimension $300 \times 2$ is generated, where $300$ represents the number of data samples and $2$ denotes the number of features. The ratio of training data and test data is set to $90\%:10\%$.
The regularization parameters $\lambda$ and $\lambda_g$ are selected from the alternatives among $\{1e^{-2}, 1e^{-4}, 1e^{-6}, 1e^{-8}, 1e^{-10}\}$ and are decayed to one-tenth every $800$ epochs. 
For each $\lambda$ ($\lambda_g$), $10$ repeated experiments are implemented with differing weight initialization.
\subsubsection{Result}
\label{appsubsubsec:Lotka-Volterra_result}
\begin{figure}
	\centering
	\begin{subfigure}[b]{0.44\textwidth}
		\centering
		\includegraphics[height = 3.7cm]{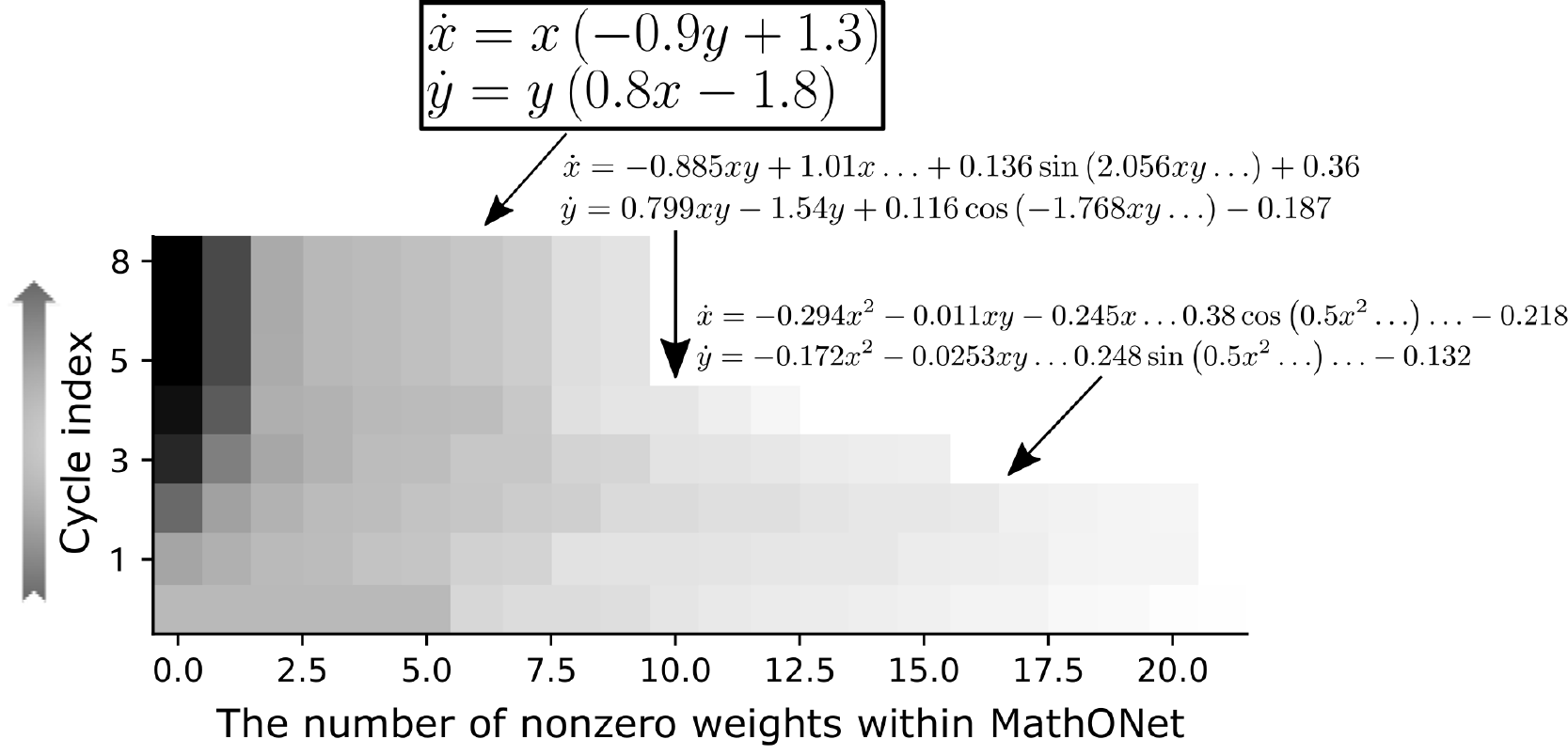}
		\caption{The number of nonzero weights of the MathONet in each cycle.}
		\label{fig:lotka_volterra_weight_equation_change}
	\end{subfigure}
	\hspace{0.1\textwidth}
	\begin{subfigure}[b]{0.43\textwidth}
		\centering
		\includegraphics[height = 4.0cm]{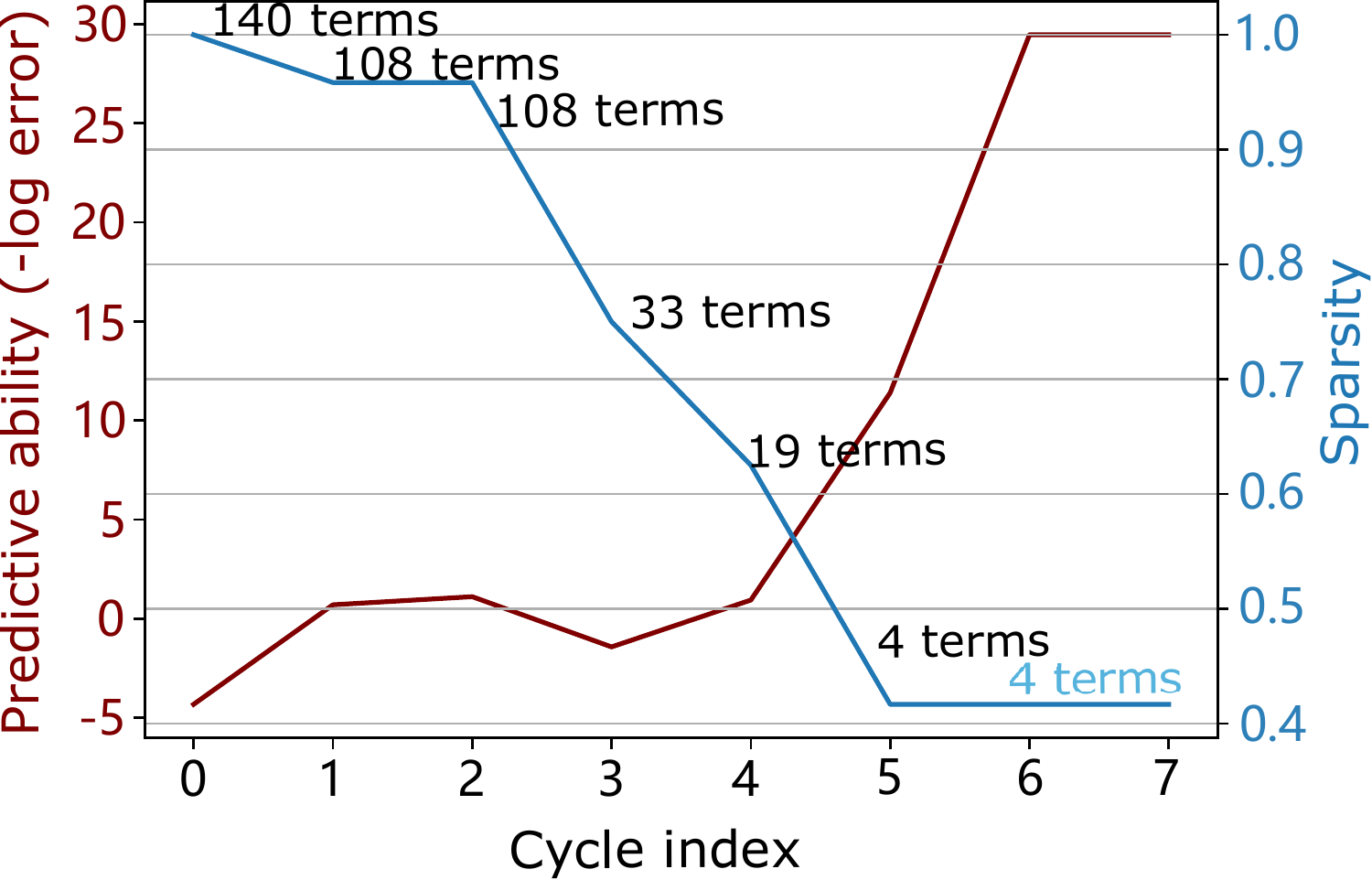}
		\caption{The sparsity and predictive ability of the MathONet in each cycle.}
		\label{fig:lotka_volterra_sparsity_loss_change}
	\end{subfigure}   
	\caption{
		The sparsity, predictive ability and weights of the identified MathONet generated in each cycle.
		a) The nonzero weights of the MathONet generated in each cycle. 
		The horizontal axis represents the combination of non-zero weights in the MathONet generated in each cycle.
		The vertical axis denotes the index of training cycles.
		The expression at each turning line (the cliff) represents the governing equation identified in the corresponding cycle.
		b) The sparsity and prediction ability of the MathONet identified in each cycle. The model becomes more and more sparse and has more and more predictive ability. The annotation next to the sparsity line represents the number of identified mathematical terms of the corresponding cycle. The first cycle represents the result identified by sparse group Lasso method which is still redundant (108 terms), and the prediction ability is low. 
	}
	\label{fig:Lotka_Volterra_weight_sparsity_loss_change}
\end{figure}

The identified model with the best prediction accuracy and being in line with Occam’s razor principle is selected as the best model. The identified system equations are:
\begin{subequations}
	\begin{align}
	&\Dot{x} = 1.300 x - 0.900 xy
	\label{eq:identified Lotka Volterra_a}
	\\
	&\Dot{y} = 0.800 xy - 1.800 y
	\label{eq:identified Lotka Volterra_b}
	\end{align}
	\label{eq:identified Lotka Volterra}
\end{subequations}
It can be observed that the identified equations are exactly the same as the theoretical system equation~\eqref{eq:theoretical Lotka Volterra}.
Fig.~\ref{fig:lotka_volterra_weight_equation_change} shows the process of the algorithm searching for the governing equations. 
Fig.~\ref{fig:lotka_volterra_sparsity_loss_change} shows the change of predictive ability and model sparsity, and the annotations along the sparsity line represent the number of retained mathematical terms of each cycle. 
It can be observed that the algorithm identify a model with only $4$ terms that originates from the initialized structure with $140$ terms.
The predictive ability improves as the model complexity decreases. 

\subsection{Fisher-KPP (Kolmogorov–Petrovsky–Piskunov) Equation}
\label{subsec:app_result_fisher}
\subsubsection{Experiment Setup}
\label{appsubsubsec:Fisher-KPP_detials}
As in~Eq.\ref{eq:theoretical fisher}, a time history of the population density $p$ and the derivative $\frac{\partial{p}}{\partial{t}}$ are used as input and output. 
The regularization parameters $\lambda$ and $\lambda_g$ are selected from the alternatives among $\{1e^{-8}, 1e^{-10}, 1e^{-12}, 1e^{-14}, 1e^{-16}\}$ and are decayed to one-tenth every $800$ epochs. 
For each hyper-parameter, the identification procedure is repeated $10$ times with differing weight initializations. 
The best model is selected according to the prediction accuracy, which is evaluated by the predicted mean square error.

\subsubsection{Result}
\label{appsubsubsec:Fisher-KPP_result}
\begin{figure}
	\centering
	\begin{subfigure}[b]{0.44\textwidth}
		\centering
		\includegraphics[height = 3.0cm]{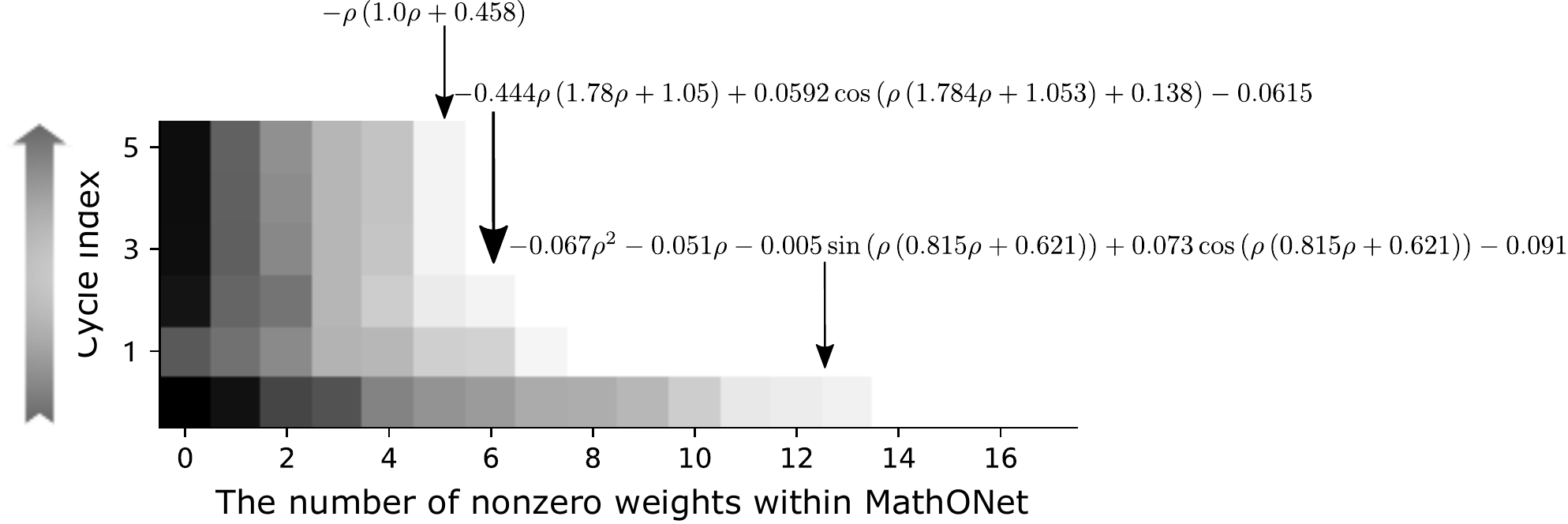}
		\caption{The number of nonzero weights of the MathONet in each cycle.}
		\label{fig:fisher_weight_equation_change}
	\end{subfigure}
	\hspace{0.1\textwidth}
	\begin{subfigure}[b]{0.44\textwidth}
		\centering
		\includegraphics[height = 3.6cm]{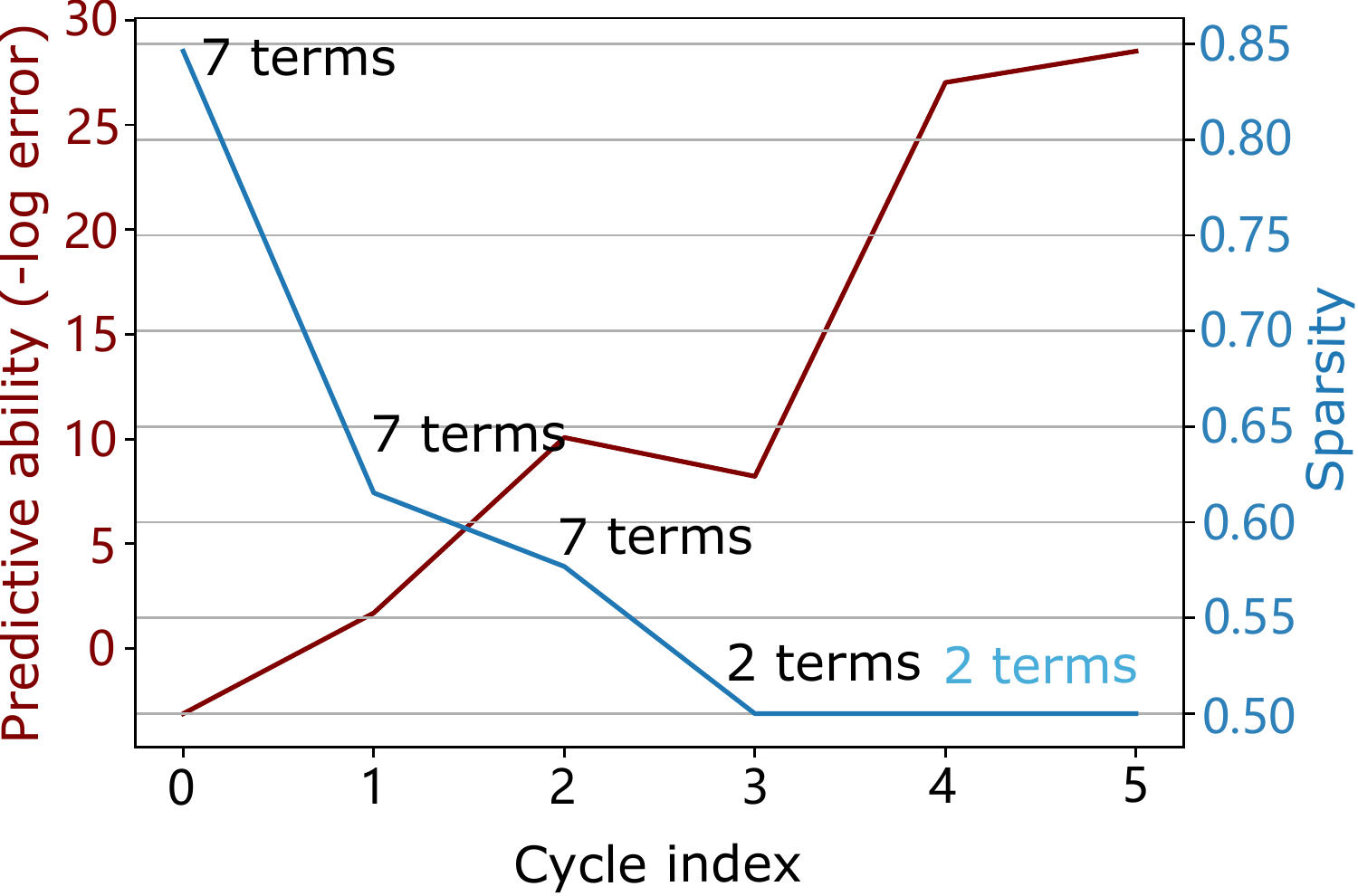}
		\caption{The sparsity and predictive ability of the MathONet in each cycle.}
		\label{fig:fisher_sparsity_loss_change}
	\end{subfigure}  
	\caption{
		The sparsity, predictive ability and weights of the identified MathONet generated in each cycle.
a) The horizontal axis represents the non-zero weights collected from the MathONet for the corresponding cycle. 
The vertical axis denotes the index of training cycles. The definition of cycle is in Algorithm~\ref{algo:algorithm}.
The expression at the cliff represents the identified governing equation of the corresponding cycle. 
b) The sparsity and prediction ability of the MathONet identified in each cycle. 
	}
	\label{fig:fisher_weight_sparsity_loss_change}
\end{figure}

Fig.~\ref{fig:fisher_weight_equation_change} shows the process of the algorithm searching for the governing equations. 
Fig.~\ref{fig:fisher_sparsity_loss_change} shows the change of predictive ability and model sparsity, and the annotations along the sparsity line represent the number of retained mathematical terms of each cycle within the MathONet.
It should be noted that MathONet contains the same number of mathematical terms ($7$ terms) in the first three cycles. However, the sparsity is gradually decreasing, which can also be observed from Fig.~\ref{fig:fisher_weight_equation_change}. This is because after some redundant connections are removed from the model, the retained connections can still represent the same or similar mathematical items represented by the redundant edges, so the number of mathematical terms remains unchanged. 
The predictive ability improves as the model complexity decreases. 

 \section{Further Discussion}
 \label{sec:discussion}
 \textbf{Regularization parameter $\lambda$ and $\lambda_g$ in Algorithm~\ref{algo:algorithm}:}
 The regularization parameters $\lambda$ and $\lambda_g$ in~\eqref{eq:loss function} needs to be properly tuned for network training.   
 With the Bayesian approach that calculates $\alpha$ and $\alpha_g$ as the determining factors for redundancy, the effort for tuning $\lambda$ can be saved a lot. 
 Typically, $\lambda$ is also the necessity for the sparse group Lasso method as shown in~\eqref{eq:conventional loss function}.
 More strict conditions are required to discover governing equations through sparse group Lasso, including proper network initialization and an extensive computational resource used for tuning $\lambda$.
 It should be noted in our algorithm, sparse group Lasso is exactly the first cycle to start and applied for all experiments. 
 However, it is challenging to discover governing equations precisely and efficiently (see Fig.~\ref{fig:lorenz_model_3_identified_arithonet_trajectory}(b) and Fig.~\ref{fig:lorenz_sparsity_loss_change_model_z} for Lorenz experiment). 




 \textbf{Comparison with deep ensemble:}
 Deep ensemble is a learning paradigm to improve generalization ability by training a set of deep neural network models with same structures and random initializations~\citep{ZHOU2002239}. 
 An ensemble includes high performing models weighted by their posterior probabilities for better accuracy and variance reduction~\citep{wilson2020case}. 
 In this work, we also train a set of MathONet models starting from random initializations using Bayesian approach. 
 However, instead of averaging on an ensemble, a single setting of parameters is selected as the optimal model by evaluating its performance.
 Although a single point mass may cause worse prediction with flawed assumptions (e.g. improper prior distribution~\citep{rasmussen2003gaussian}), it also alleviates the issue of a tremendous computational expense that deep ensembles may incur. 












 \textbf{Comparison with variational inference:}
 Variational inference (VI) is another typical approximation method for Bayesian inference by minimizing the Kullback-Leibler divergence between an assumed approximated posterior and true posterior distribution.
 VI method can provide bounds on probabilities of interest and yield deterministic approximation procedures without tuning regularization parameter~\citep{jordan1999introduction}. Its applications on model compression have been explicitly interpreted in~\citep{hinton1993keeping,louizos2017bayesian} 
 However, the manual selection for proper pruning thresholds is also required, which hinder its compression efficiency for complex models.
 In contrast, the Laplace approximation method can be implemented more efficiently and extended to complex models. 
 It is also worth studying to extend the sparse group Bayesian approach to network compression by enforcing various structural sparsity over network parameters.

\end{document}